\documentclass[letterpaper]{article} % DO NOT CHANGE THIS
\usepackage{aaai25}  % DO NOT CHANGE THIS
\usepackage{times}  % DO NOT CHANGE THIS
\usepackage{helvet}  % DO NOT CHANGE THIS
\usepackage{courier}  % DO NOT CHANGE THIS
\usepackage[hyphens]{url}  % DO NOT CHANGE THIS
\usepackage{graphicx} % DO NOT CHANGE THIS
\usepackage{natbib}  % DO NOT CHANGE THIS AND DO NOT ADD ANY OPTIONS TO IT
\usepackage{caption} % DO NOT CHANGE THIS AND DO NOT ADD ANY OPTIONS TO IT
\usepackage{algorithm}
\usepackage{booktabs}
\usepackage{graphicx}
\usepackage{mathtools}

\usepackage{amsmath, amssymb, amsfonts}
\usepackage{dsfont}
\usepackage{textcomp}
\usepackage{comment}
\usepackage{subcaption}
\usepackage{multirow}
\usepackage{makecell}
\usepackage{tabularx}
\usepackage{xspace}
\usepackage{xcolor}
\usepackage{stmaryrd}
\usepackage{booktabs}
\usepackage{algpseudocode}
\usepackage{amsmath}
\usepackage{cleveref}
\usepackage{adjustbox}

\usepackage{colortbl}
\usepackage{hhline}

\usepackage{newfloat}
\usepackage{listings}
\DeclareCaptionStyle{ruled}{labelfont=normalfont,labelsep=colon,strut=off} % DO NOT CHANGE THIS
\floatstyle{ruled}
\newfloat{listing}{tb}{lst}{}
\floatname{listing}{Listing}
\usepackage{amsmath}
\newcommand{\bfsection}[1]{\noindent\textbf{#1}.}

\newcommand{\modelnamelong}{Multimodal ASsociation Score\xspace}
\newcommand{\modelname}{MASS\xspace}

\title{\modelname: Overcoming Language Bias in Image-Text Matching}

\author{
    Jiwan Chung\textsuperscript{\rm 1},
    Seungwon Lim\textsuperscript{\rm 1},
    Sangkyu Lee\textsuperscript{\rm 1},
    Youngjae Yu\textsuperscript{\rm 1}
}
\affiliations{
    \textsuperscript{\rm 1}Yonsei University\\
    50 Yonsei-ro, Seodaemun-gu\\
    Seoul, South Korea\\
    jiwan.chung.research@gmail.com
}

\usepackage{bibentry}
\newcommand{\eg}{{\it e.g.,}}%
\newcommand{\ie}{{\it i.e.,}}%
\newcommand{\etal}{{\it et al.}}%

\begin{document}

\maketitle

\begin{abstract}
Pretrained visual-language models have made significant advancements in multimodal tasks, including image-text retrieval.
However, a major challenge in image-text matching lies in language bias, where models predominantly rely on language priors and neglect to adequately consider the visual content.
We thus present \modelnamelong (\modelname), a framework that reduces the reliance on language priors
for better visual accuracy in image-text matching problems.
It can be seamlessly incorporated into existing visual-language models without necessitating additional training. Our experiments have shown that \modelname effectively lessens language bias %, thereby improving the performance of visual-compositionality understanding tasks.
without losing an understanding of linguistic compositionality.
Overall, \modelname offers a promising solution for enhancing image-text matching performance in visual-language models.
\end{abstract}

\section{Introduction}
\label{sec:intro}

Recently, pretrained visual-language models have showcased impressive performance across a broad spectrum of multimodal tasks~\cite{li2022blip,alayrac2022flamingo,wang2022ofa}, including simple image-text retrieval~\cite{chen2015microsoft,young2014image}. 
In particular, CLIP~\cite{Radford2021LearningTV} has emerged as one of the most popular models for image-text matching.
The image-text similarity score derived from CLIP is shown to be effective for assessing image caption quality~\cite{hessel2021clipscore},
providing reward signals to train multimodal models~\cite{yu2023fusing},
and building a large-scale image search system~\cite{beaumont2022clip}.

However, CLIP falters when the given task requires accurate language modeling.
Recent benchmarks have unveiled CLIP's deficiency in modeling
linguistic compositionality
~\cite{thrush2022winoground, nikolaus2022vision,yuksekgonul2022and}.
Furthermore,
current visual-language models also have demonstrated inconsistent outcomes on image-text matching that demand comprehension of linguistic constructs,
such as the existence and quantity of objects or coreference~\cite{parcalabescu2022valse,shekhar2017foil}.

\begin{figure}
    \centering
    \includegraphics[width=0.48\textwidth]{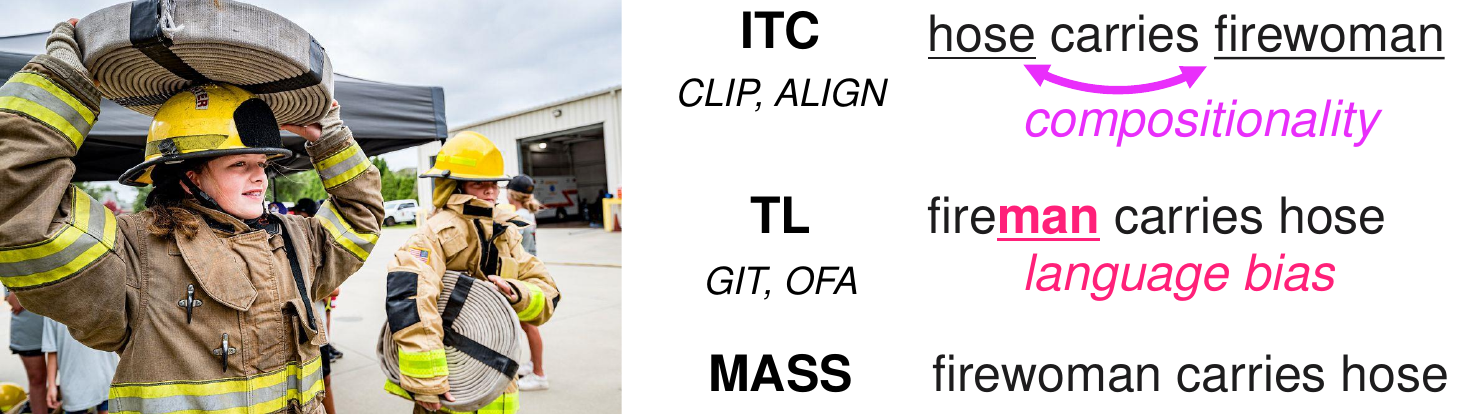}
    \caption{
    Captions retrieved with each method given the image, where only \modelname succeeds in ruling out the failure modes.
    Models trained Image-Text Contrastive (ITC) objectives such as CLIP~\cite{Radford2021LearningTV} 
    fail to model linguistic structure. % and assigns higher scores for captions with wrong prepositions.
    Token Likelihood (TL) of image captioning models including OFA~\cite{wang2022ofa} shows overreliance on its language prior.
    Our \modelname amends the language bias of image captioning models for accurate image-text matching capability.
    }
    %\caption{The left image is aligned to the caption \textit{a person stands and a dog sits}, while the right image is not. Our \modelname looks at different parts of the text (\textit{stands} vs. \textit{dog}) depending on the given image, unlike the token likelihood $p(text|image)$ which relies on language bias to show similar likelihood patterns for both images.}
    \label{fig:intuition}
\end{figure}

\begin{figure*}[ht]
    \centering
    \includegraphics[width=0.98\textwidth]{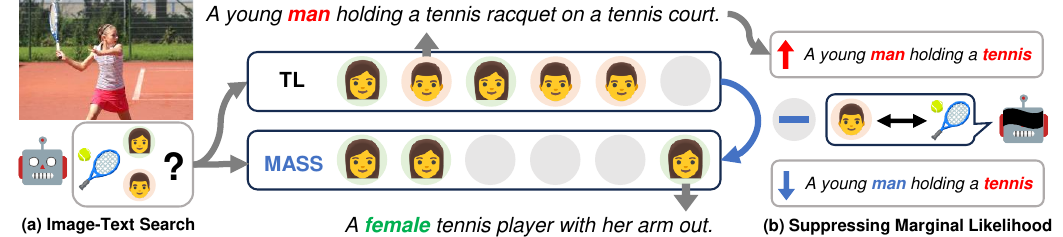}
    \caption{
    (a) Given an image of a girl playing tennis, the visual-language model falsely retrieves captions describing the subject as male
    by relying on language bias. (b) On the other hand, \modelname reduces such gender bias by adopting pointwise mutual information which suppresses the text-only marginal likelihood.
    We provide the corresponding experimental results
    %in~\labelcref{subsec:exp_gender}
    in~\cref{subsec:exp_gender}.
    }
    \label{fig:arch}
\end{figure*}

As an alternative to the contrastive objective in CLIP, recent studies~\cite{petryk2024simple,tschannen2024image} revealed that
log-likelihood induced from
the autoregressive image captioning objective
features a better understanding of linguistic structures,  
including compositionality.
The objective 
fosters language generation capability
in the trained models, 
inducing stronger linguistic understanding.

Still, there is a caveat in directly using the image captioning models
to assess image-text similarity.
While training for the captioning objective,
the models inherently build a prior for language distribution
on what text sequence is more likely
\textit{regardless of the given image}.
As shown in Figure~\ref{fig:intuition},
overreliance on this prior
leads to weaker visual conditioning
and then to incorrect image-text matching.
% One contributing factor to this subpar performance is language bias.
Drawing inspiration from previous research~\cite{Niu2020CounterfactualVA}, we refer to such phenomenon as
\textit{language bias}.
Language bias refers to the propensity of visual-language models to rely heavily on language priors in the training data instead of properly conditioning their output on the given images.
It has been repeatedly reported that language bias is a major bottleneck in visual question-answering~\cite{agrawal2016analyzing,zhang2016yin,agrawal2018don, si2022towards}, and other vision-language tasks~\cite{srinivasan2021worst,salin2022vision}.

%Then how does language bias influence image-text matching? %To illustrate this, 
%To demonstrate the detrimental impact of language bias on image-text matching,
%we reference the first row of Figure~\ref{fig:intuition}, where two images and the likelihood of text $p(text|image)$ per token from an image captioning model~\cite{wang2022ofa} are depicted.
%Here, the model shows similar patterns of likelihood, regardless of the given images, indicating a reliance on language bias. As a result, when aggregating over the token likelihood, the model would in turn falsely assign similar scores to both image-text pairs.
%Also, CLIP~\cite{radford2021learning} and OFA~\cite{wang2022ofa} are shown to similarly suffer from language bias in our bias reduction experiments (\cref{sec:exp_bias}) which are formulated as image-text matching problems.

In this paper, we present \modelnamelong (\modelname) as an inference-time framework designed to reduce language bias in image-text matching.
\modelname aims to measure the strength of the association
between image and text modality
without involving the textual prior.
% Inspired by pointwise mutual information, a technique widely used to measure the association between two variables in domains including data mining~\cite{davison2019commonsense} or multi-choice QA~\cite{holtzman2021surface}.
Our approach uniquely functions as follows:
Given a pretrained visual-language model,
we first extract the image-conditional and text-only likelihood per text token. We then compute the pointwise mutual information of each image and text token and aggregate them for the text sequence to create a debiased scalar similarity score.
Importantly, \modelname can be computed for any off-the-shelf visual-language model that outputs image-conditioned text likelihood.
Also, it does not require any additional training.

We first demonstrate that \modelname can mitigate language bias in visual-language models
by evaluating it with multimodal color~(\cref{subsec:exp_color}), number~(\cref{subsec:exp_number}), and gender~(\cref{subsec:exp_gender}) bias benchmarks.
Our results indicate that \modelname significantly improves performance over CLIP~\cite{Radford2021LearningTV} and raw token likelihood (TL) in the datasets with clear language bias (color), explicit bias in labels (number), and practical image search problems (gender) about reliance on language bias. 
% Interestingly, \modelname's ability to mitigate language bias is also shown to alleviate societal (gender) bias.
We further evaluate \modelname's capability to maintain linguistic understanding against two visual-linguistic compositionality benchmarks;
Winoground~\cite{thrush2022winoground} and SVO-Probe~\cite{hendricks2021probing}.
\modelname greatly enhances the performance of the backbone visual-language model in both experiments and outperforms the strong baselines.

The main contributions of this paper are threefold.
% Firstly, our inference framework is the first to apply pointwise mutual information in reducing language bias of visual-language models.
% Secondly, we demonstrate that the language bias reduction facilitated by our \modelname leads to better performance in visual-linguistic understanding benchmarks.
Firstly, we present \modelname as an effective framework for language bias reduction in image-text matching tasks.
Secondly, we demonstrate that \modelname can both reduce language bias and maintain linguistic understanding with various biases and compositionality benchmarks.
Lastly, our approach does not necessitate additional training and can be directly applied to a wide variety of visual-language models for robust image-text matching.

\section{Image-Text Similarity Functions}
\label{sec:method_pre}

% Our goal is to find a similarity function that correctly matches image and text in the presence of linguistic complexities such as visual-linguistic compositionality.
Our objective is to identify a similarity function that accurately pairs image and text, even in the presence of language bias in the pretrained model.
Given an image $\mathbf{c} \in \mathcal{C}$ and a text $\mathbf{x} \in \mathcal{X}$, the similarity function $\mathcal{S}(\mathbf{c}, \mathbf{x})$ evaluates the closeness of alignment such that parallel image-text pairs have high scores.
We divide the similarity functions by the granularity of the training objectives:
Image-Text Contrastive Learning (ITC) and Image-Text Matching (ITM) provide scores on \textit{sequence-level}~(\cref{subsec:method_seq}), while Token-Level supervision (TL) such as supervised image captioning applies \textit{token-level}~(\cref{subsec:method_tok}) feedback.

\subsection{Sequence-Level Similarity Functions}
\label{subsec:method_seq}

The sequence-level similarity functions are defined on top of dual vision-language encoders. The dual encoders may be either \textit{separate} (\ie ~architectures with independent encoders per modality as CLIP~\cite{Radford2021LearningTV} and the lower modules of ALBEF~\cite{li2021align}) or \textit{intertwined} (\eg~the upper module of ALBEF) in terms of architecture.

\bfsection{Image-Text Contrastive Learning}
ITC is a training objective for learning multimodal joint embedding space based on the Noise Contrastive Estimation~\cite{oord2018representation}.
%by maximizing mutual information between image $\mathbf{c}$ and text $\mathbf{x}$. 
We use \textit{separate} image and text encoders $f_{\phi}$, $g_{\psi}$
%, and one-side objective for explanation here, which is a go-to architecture for models and typically trained but symmetric ITC objective~\cite{radford2021learning}.
as often used with the ITC objective~\cite{Radford2021LearningTV}.

ITC requires the aligned image and text pair $(\mathbf{c}, \mathbf{x})$ and a set of misaligned text $\mathbf{\widetilde{x}}$ typically sourced from other text within the minibatch.
ITC first obtains vector embeddings of each image and text and computes the cosine similarity of them to build image-text pair logits,
which are then used to calculate the cross-entropy loss.
%We can calculate image-text similarity as the similarity between embedding from dual encoder with trained parameters ($\bar{\phi}, \bar{\psi}$), and the most popular choice of similarity function is cosine similarity~\cite{radford2021learning, hessel2021clipscore}.
We use the term ITC both to denote the training objective and corresponding similarity function $\mathcal{S}_{\text{ITC}}$ induced from the trained model with parameters $\bar{\psi}$ and $\bar{\phi}$.
In experiments, we include CLIP~\cite{Radford2021LearningTV} as the baseline
similarity function of this type.

\begin{align}
    \mathcal{S}_{\text{ITC}}(\mathbf{c}, \mathbf{x}) 
    &\coloneqq 
    \cfrac{ f_{\bar{\phi}}(\mathbf{c})^{T} \cdot g_{\bar{\psi}}(\mathbf{x}) }{|| f_{\bar{\phi}}(\mathbf{c}) || \cdot ||g_{\bar{\psi}}(\mathbf{x})|| }
\end{align}

% The inner product between embedding on each encoder acts as a critic function proportional to the ratio between joint density $p(\mathbf{c}, \mathbf{x})$ and the product of each marginal density $p(\mathbf{c})$ and $p(\mathbf{x})$, which is trained to maximize on the aligned text pair by cross entropy loss.
% \begin{align}
%     \mathcal{L}_{\text{ITC}}(\mathbf{c}, \mathbf{x}) = -\cfrac{\exp (f_{\phi}(\mathbf{c})^{T} \cdot g_{\psi}(\mathbf{x}))}{\sum_{\mathbf{\widehat{x}} \in \{\mathbf{x}, \mathbf{\widetilde{x}}\}} \exp (f_{\phi}(\mathbf{c})^{T} \cdot g_{\psi}(\mathbf{\mathbf{\widehat{x}}}))}
%     % &\coloneqq 
%     % \cfrac{ g_{\bar{\phi}}(\mathbf{c})^{T} \cdot h_{\bar{\psi}}(\mathbf{x}) }{|| g_{\bar{\phi}}(\mathbf{c}) || \cdot ||h_{\bar{\psi}}(\mathbf{x})|| }
% \end{align}

\bfsection{Image-Text Matching}
ITM is a binary classification objective
that decides whether a pair of image and text $(\mathbf{c}, \mathbf{x})$ is aligned or not
%in vision-language pretraining literature
~\cite{li2021align, wang2022git, wang2022ofa, li2022blip}.
While both ITM and ITC require negative samples for training, 
ITC typically compares a positive sample with a large number of negatives,
while ITM determines the correctness of a single sample.
In practice, ITM is often used to train \textit{intertwined} models~\cite{li2021align, li2022blip}, which we adopt for explanation here.

For each pair of images and text $(\mathbf{c}, \mathbf{x})$,
an \textit{intertwined} image-text encoder $f_{\phi, \psi}$ processes both to yield a multimodal vector representation. Then, a linear classifier $h_{\omega}$ is applied on top of the representation to build
%logit and trained by binary cross entropy loss. 
a binomial logit for the cross-entropy loss.
At inference time,
%logit with sigmoid function from
the probability of the image-text pair being true
%trained parameter ($\bar{\phi}, \bar{\psi}, \bar{\omega}$)
is used as the similarity function output $\mathcal{S}_{\text{ITM}} (\mathbf{c}, \mathbf{x})$.
In our setting, all baselines marked ITM fall into this category.
Note that OFA uses the text generation head to obtain the ITM score instead of a separate linear classifier. Refer to the corresponding paper~\cite{wang2022ofa} for details.
\begin{align}
    \mathcal{S}_{\text{ITM}}(\mathbf{c}, \mathbf{x}) = \cfrac{\exp ( h_{\bar{\omega}} ( f_{\bar{\phi}, \bar{\psi}} (\mathbf{c}, \mathbf{x}) ) )}{1 + \exp ( h_{\bar{\omega}} ( f_{\bar{\phi}, \bar{\psi}} (\mathbf{c}, \mathbf{x}) ) )}
\end{align}

\subsection{Token-Level Similarity Functions}
\label{subsec:method_tok}

\bfsection{Token-Level Likelihood}
Let us divide a sample of text into a sequence of \textit{tokens} $\mathbf{x} = \{x_1, \cdots, x_l\}$ of length $l$.
The token-level functions require models to output the image-text alignment score for each text token.
A popular training objective that admits such constraint is image captioning.
Given a pair of image-text $(\mathbf{c}, \mathbf{x})$, an image captioning model with parameter $\theta$ is trained to maximize the log-likelihoods $\log p_{\theta}$ of the text sequence $\mathbf{x}$ factorized in an autoregressive manner. 
With trained parameter $\bar{\theta}$, the model likelihood for the probability of the next token conditioned on both the image and previously generated text $p_{\bar{\theta}} (x_t | x_{<t}, \mathbf{c})$ can be treated as an approximation to the true conditional likelihood $p (x_t | x_{<t}, \mathbf{c})$.

% \begin{align}
%     \mathcal{L}_{\text{TL}} (\mathbf{c}, \mathbf{x})
%     &= - \frac{1}{l} \sum_{t \le l} \log p_{\theta} (x_t | x_{<t}, \mathbf{c})
% \end{align}

\bfsection{Aggregation}
The token-level likelihood of an image captioning model can be directly used as the token-level similarity function~\cite{petryk2024simple}. For all token-level similarity scores, we compute the mean of all text token scores in the sequence to build the sequence-level image-text alignment metric. We denote this type of similarity function with TL in the experiments.
\begin{align}
    \mathcal{S_{\text{TL}}}(\mathbf{c},\mathbf{x}) 
    &\coloneqq
    \cfrac{1}{l} \sum_{t < l} p_{\bar{\theta}} (x_t | x_{<t}, \mathbf{c})
\end{align}

\section{\modelnamelong}
\label{sec:method_ours}

Directly using likelihood from autoregressive visual-language models as an image-text similarity function is problematic since this does not represent \textit{pure} image-text similarity. 
For instance, image captioning models are also required to generate \textit{linguistically plausible} captions. 
Whether the token $x_t$ matches the given image $\mathbf{c}$ or not, the model can assign a high likelihood if it was exposed frequently with the prefix $x_{<t}$ in the training process. To address this ignorance of the context, \textit{pointwise mutual information} $PMI(\mathbf{x}; \mathbf{c})$ has been utilized in NLP tasks such as the conversation model~\cite{li2015diversity} and text summarization~\cite{van2022mutual}.
% \modelname is the first attempt at applying PMI to autoregressive visual-language models, and we empirically show its effectiveness for reducing language bias in image-text matching.
However, its application to the visual-language model has yet to be investigated, and we propose it as an effective method for reducing language bias in image-text matching.
\begin{align}
    PMI(\mathbf{x}; \mathbf{c}) \coloneqq \log \cfrac{p(\mathbf{x}, \mathbf{c})}{p(\mathbf{x}) p(\mathbf{c})} = \log \cfrac{p(\mathbf{x}|\mathbf{c})}{p(\mathbf{x})}
\end{align}

\subsection{Language Bias Reduction}\label{subsec:method_debiasing}

% Thus, we formulate how pointwise mutual information can debias the language bias of the autoregressive visual-language model (\cref{subsec:method_debiasing}) and how can we efficiently calculate pointwise mutual information as similarity function in token-level (\cref{subsec:method_marginal}).

Formally, consider the log-likelihood of a given image and text pair $(\mathbf{c}, \mathbf{x})$ from trained parameter $\bar{\theta}$, processed in an autoregressive manner. According to Bayes' rule, we can decompose the log-likelihood to two distinct terms based on whether image $\mathbf{c}$ is related to the text $\mathbf{x}$ or not. We label the term not related to the $\mathbf{c}$ as the \textit{linguistic} log-likelihood and the term that is related to the $\mathbf{c}$ as the \textit{association} log-likelihood.
\begin{align}
    \log p_{\bar{\theta}}(\mathbf{x}|\mathbf{c}) 
    = \underbrace{ \vphantom{\cfrac{p_{\bar{\theta}}(\mathbf{c}|\mathbf{x})}{p_{\bar{\theta}}(\mathbf{c})}} \log p_{\bar{\theta}}(\mathbf{x}) }_{\text{linguistic}} 
    + \underbrace{ \log \cfrac{p_{\bar{\theta}}(\mathbf{c}|\mathbf{x})}{p_{\bar{\theta}}(\mathbf{c})} }_{\text{association}}
\end{align}

Under this perspective, the linguistic log-likelihood represents the linguistic plausibility of text $\mathbf{x}$.
While the term is independent of the image context,
it can overrule the similarity between image $\mathbf{c}$ and text $\mathbf{x}$ measured by the \textit{association} term when the model relies on text bias.
% This is an \textit{independent} property,  the similarity between image $\mathbf{c}$ and text $\mathbf{x}$, but it can amplify the log-likelihood even when the correspondence log-likelihood is negligible. 
As such, a simple remedy for the likelihood-based similarity functions would be to regularize the log-likelihood by subtracting the linguistic log-likelihood. 
For the token-level similarity function, we can deduct the \textit{token-level} linguistic log-likelihood $\log p_{\bar{\theta}} (x_t | x_{<t})$ from token-level log-likelihood $\log p_{\bar{\theta}} (x_t | x_{<t}, \mathbf{c})$ for each token $x_t$. 

It is noteworthy that if $p_{\bar{\theta}} (\mathbf{x})$ and $p_{\bar{\theta}} (\mathbf{x} | \mathbf{c})$ are accurately estimating the ground-truth marginal density $p(\mathbf{x})$ and conditional density $p(\mathbf{x}|\mathbf{c})$, this approach equates to estimating the pointwise mutual information. 
In this sense, we denote this similarity function as \modelnamelong (\modelname), which is calculated by averaging each token's pointwise mutual information over the total length $l$. 
\begin{align}
    \mathcal{S_{\text{\modelname}}}(\mathbf{c},\mathbf{x}) 
    &\coloneqq
    \cfrac{1}{l} \sum_{t < l} \log \cfrac{p_{\bar{\theta}} (x_t | x_{<t}, \mathbf{c})}{p_{\bar{\theta}} (x_t | x_{<t})}
\end{align}

% Note that the target quantity with exponential function is proportional to the critic function of dual-encoder trained with contrastive learning used in ITS score. The dual-encoder architecture does not guarantee the critic function to be normalized, due to the energy-based modeling for Noise Constrastive Estimation (NCE) framework. This property made direct comparison of the scale of the critic function hard at the test time, so surrogate similarity measure such as cosine similarity should be used for score estimator. 

% However, the autoregressive architecture decomposed the density $p(\mathbf{y}|\mathbf{x})$ to the tractable chain of token probability $p(\mathbf{y}_i|\mathbf{x}, \mathbf{y}_{<i})$, we can directly compare the pointwise estimation of target quantity as the score of the confidence if we can successfully estimate the marginal probability $p(\mathbf{y})$ by $f_{\theta}(\mathbf{y})$. 

\subsection{Estimating the Marginal Likelihood}\label{subsec:method_marginal}

The autoregressive visual-language model inherently assumes an image input $\mathbf{c}$, as it is trained to estimate the \textit{conditional} log-likelihood $\log p(\mathbf{x}|\mathbf{c})$. As a result, we cannot directly obtain the token-level marginal log-likelihood $\log p_{\bar{\theta}}(x_t|x_{<t})$, which is required to calculate the \modelname score. One possible solution is to use a Monte Carlo approximation by uniformly sampling $N$ random image $\widetilde{\mathbf{c}}$. For each token $x_t$, we can approximate the marginal log-likelihood $\log p_{\bar{\theta}}(x_t|x_{<t})$ by taking the average of the conditional log-likelihood $\log p_{\bar{\theta}}(x_t|x_{<t}, \widetilde{\mathbf{c}})$ over random images $\widetilde{\mathbf{c}}_i$.
\begin{align}
    \log p_{\bar{\theta}}(x_t|x_{<t}) &= \int_{\mathcal{C}} \log p_{\bar{\theta}}(x_t|x_{<t}, \mathbf{c}) d\mathbf{c} \\
    &\approx \cfrac{1}{N}\sum_{i=1}^N \log p_{\bar{\theta}}(x_t|x_{<t}, \widetilde{\mathbf{c}}_i)
\end{align}

However, this approach requires $N$ more forward passes to compute the score for one image-text pair, and it is also challenging to ensure the accuracy of the approximation. As an alternative, we discovered that using image input as a \textit{null image} $\mathbf{c}_{\emptyset}$, which is a 'black-filled' image, is a good alternative for estimating $\log p_{\bar{\theta}}(x_t|x_{<t})$. For the rest of the paper, we use $\log p_{\bar{\theta}}(x_t | x_{<t}, \mathbf{c}_{\emptyset})$ as the approximation of $\log p_{\bar{\theta}}(x_t | x_{<t})$ to calculate \modelname. 
\begin{align}
    \mathcal{S_{\text{\modelname}}}(\mathbf{c},\mathbf{x}) 
    &\approx
    \cfrac{1}{l} \sum_{t < l} \log \cfrac{p_{\bar{\theta}} (x_t | x_{<t}, \mathbf{c})}{p_{\bar{\theta}} (x_t | x_{<t}, \mathbf{c}_{\emptyset})}
\end{align}
\section{Experiments: Language Debiasing}
\label{sec:exp_bias}

We show that \modelname effectively reduces language bias in three different domains;
natural color association (\cref{subsec:exp_color}), object number counting (\cref{subsec:exp_number}), and gender balance in image search (\cref{subsec:exp_gender}).
The improvement is especially substantial when
the tested data have a different language bias from the training data
of the base visual-language model.

\bfsection{Backbones}
We use OFA~\cite{wang2022ofa} as the backbone VL model for both ITM, TL, and \modelname similarity functions.
Also, we experiment with alternative backbones of BLIP-2~\cite{li2023blip} and LLaVA~\cite{liu2024visual}. Note that ITM is not applicable for LLaVA since it only has the language modeling head.

% \begin{figure}[t]
%     \centering
%     \includegraphics[width=0.48\textwidth]{figure/fig_NCD_color_bias.pdf}
%     \caption{Description about how image-text pair is given to model. We compute difference between the score of 'fruit image-true caption' and 'fruit image-adversarial caption'}
%     \label{fig:fruit_gray}
% \end{figure}

%\begin{figure}[t]
%    \includegraphics[width=0.48\textwidth]{figure/fig_color.pdf}
%    \caption{Colour bias experiments on the Natural Colors Dataset~\cite{anwar2020image}. X-axis represents score differences which should be greater than zero.}%  x :  unmatching pair score - matching pair score, y: fruit type of NCD. large difference means less-biased for fruit color.}
%    \label{fig:exp_color}
%\end{figure}

% We find that our method's success on Imagescore (and groupscore because Imagescore is a bottleneck for groupscore on most of VL model and most of scoring method) is attributed to debiasing ability of our scoring scheme in language aspect. Table 6 shows VL model can concentrate more on objects in image, \\
% 
% 
% To estimate the capability of debiasing language prior inevitably gained through training process, we conducted 2 experiments that can prove language bias can be reduced by our method. We validate the hypotheses by two different experiment, colour debiasing and social(especially gender) debiasing.\\[10pt]

\begin{figure}[t]
    \centering
    \begin{subfigure}[b]{0.48\textwidth}
        \centering
        \includegraphics[width=\textwidth] {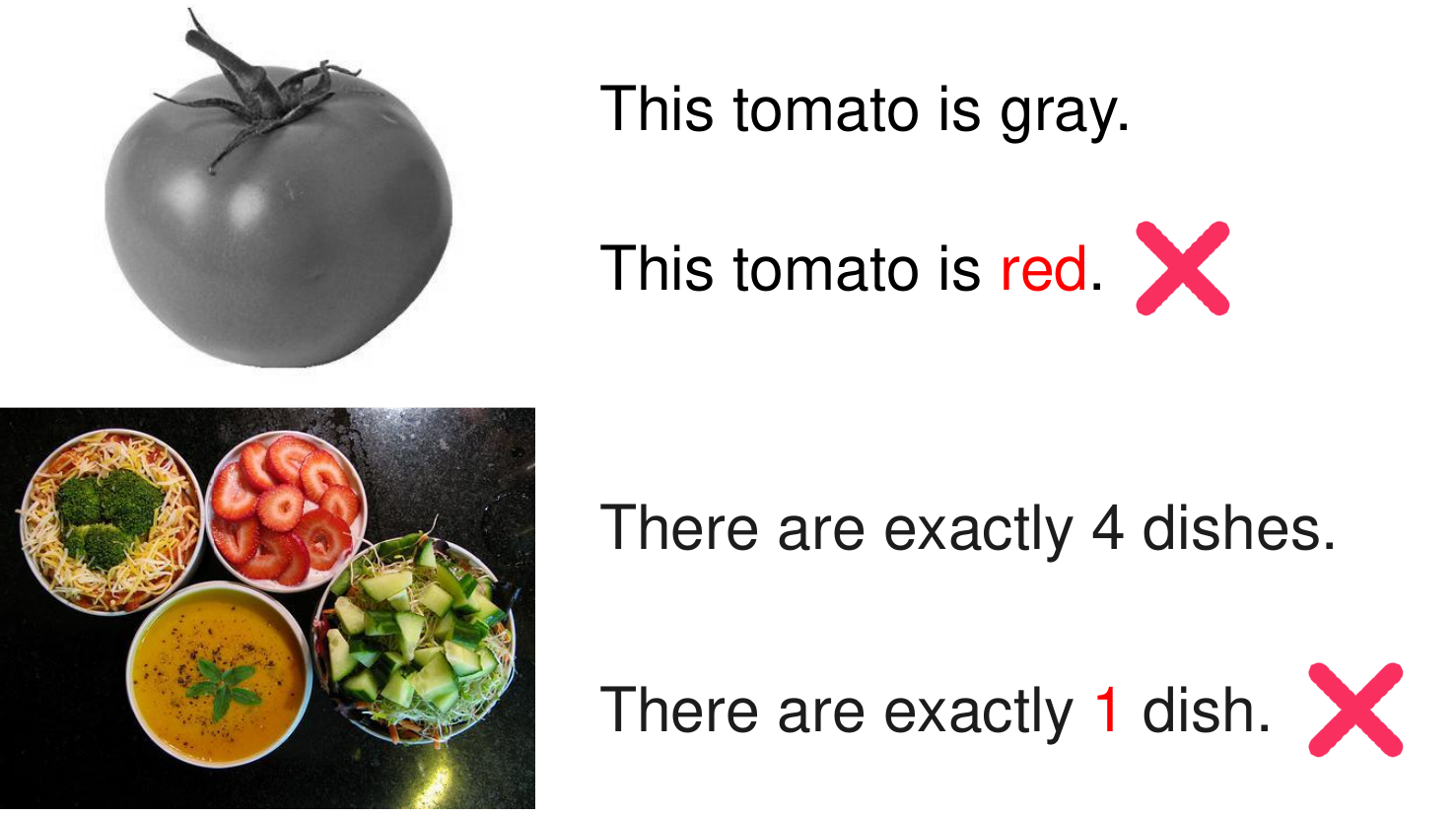}
        \caption{Top: A sample from our color debiasing experiment. A model without language bias would say the tomato is gray. Bottom: A sample from the counting benchmark.
        %In the adversarial subset, the correct answer is always bigger than three while models typically prefer smaller numbers~\cite{Parcalabescu2021VALSEAT}.}
        }
        \label{fig:bias_examples}
    \end{subfigure}
    \hfill
    \begin{subfigure}[b]{0.48\textwidth}
        \centering
        \includegraphics[width=\textwidth] {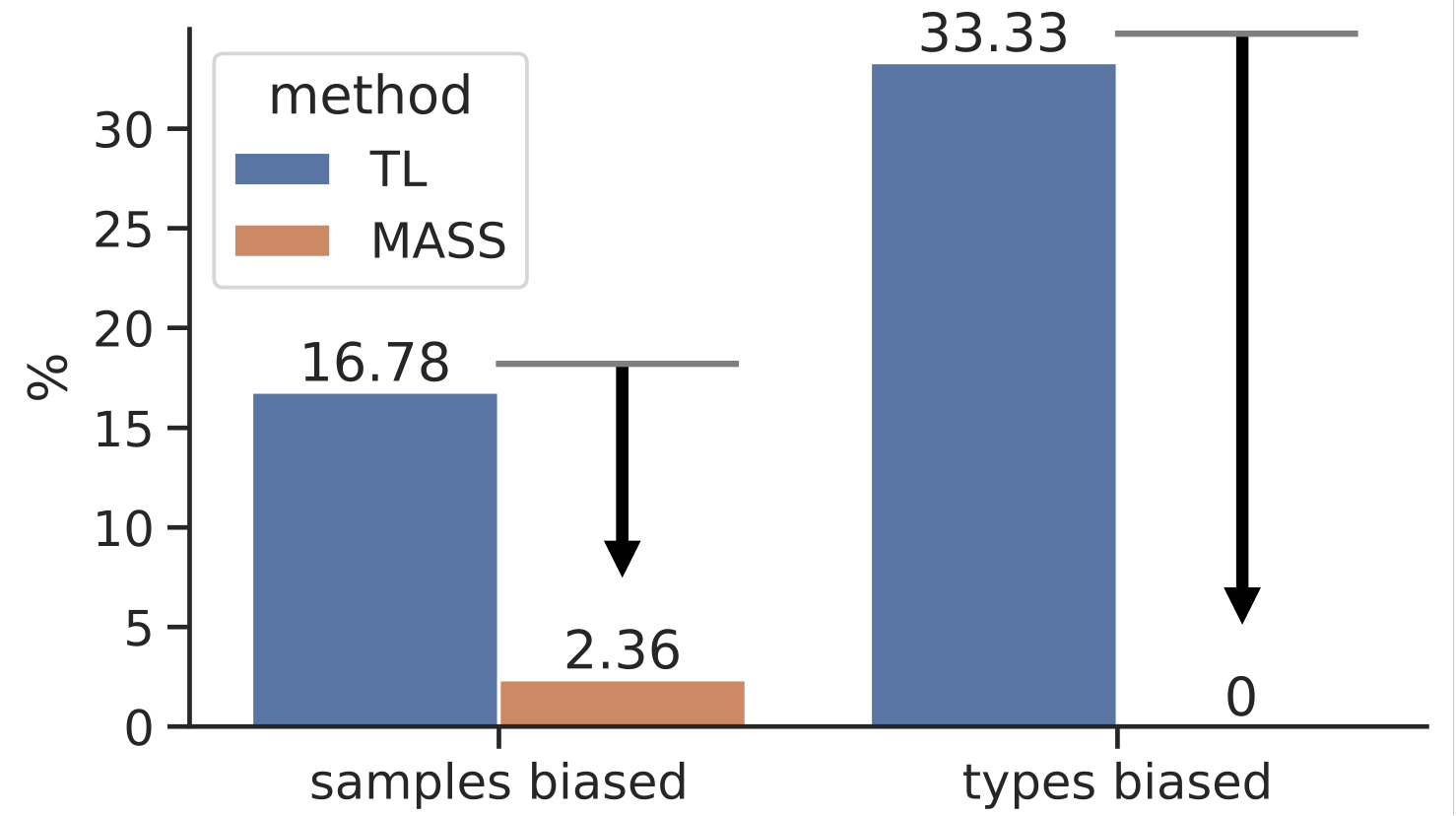}
        \caption{Results on Natural Colors Dataset~\cite{anwar2020image}. \modelname effectively debiases TL both in sample and type-level.}
        \label{fig:color_bias}
    \end{subfigure}
    \caption{Data samples and experimental results from our color debiasing experiment.}
    \label{fig:color_dual}
\end{figure}
\begin{table*}
    \centering
    \begin{tabular}{l|ccc|ccc|ccc}
         & \multicolumn{3}{c}{balanced} & \multicolumn{3}{c}{Small} & \multicolumn{3}{c}{Adv.} \\
         Models
        & ITM/C & TL & \textbf{MASS}
        & ITM/C & TL & \textbf{MASS}
        & ITM/C & TL & \textbf{MASS}
        \\
        \hline
        Chance & \multicolumn{3}{c}{16.7} & \multicolumn{3}{c}{16.7} & \multicolumn{3}{c}{16.7} \\
        \hline
        CLIP & 62.1 & - & - & 62.5 & - & - & 57.5 & - & - \\
        LXMERT & 62.2 & - & - & 69.2 & - & - & 42.6 & - & - \\
        12-in-1 & 76.7 & - & - & 80.2 & - & - & 77.3 & - & - \\
        \hline
        OFA\textsubscript{large}
        & 66.8 & 65.4 & \textbf{70.0} ($\uparrow$3.2)
        & 71.7 & 71.1 & \textbf{73.2} ($\uparrow$1.5)
        & 72.0 & 66.8 & \textbf{76.7} ($\uparrow$4.7)\\
        \hline
        % BLIP-2\textsubscript{T5}
        %& 68.0 & 59.3 & 63.6
        %& 69.8 & 66.1 & 67.4
        %& 77.6 & 58.6 & 72.0 \\
        %\xspace\xspace\xspace\xspace\xspace\xspace\xspace\xspace\xspace\xspace\textsubscript{OPT}
        %& 68.0 & 63.3 & 58.3
        %& 69.8 & 65.6 & 63.3
        %& 77.6 & 62.2 & 70.2 \\
        %\hline
        
    \end{tabular}
    \caption{Counting experiment in VALSE dataset~\cite{parcalabescu2022valse}. Adv. denotes the adversarial set with explicit language bias. The best numbers given the same backbone are \textbf{bolded}.}
    \label{tab:count}
\end{table*}

\subsection{Debiasing Colors}
\label{subsec:exp_color}

When there is linguistic bias in the data, the visual-language models trained on them would likely reflect the bias~\cite{birhane2021multimodal,ross2021measuring}.
However, evaluating the model for bias is not a simple problem:
how do we know when the model only relies on the text context without looking at the image?

Here, we devise a toy experiment where the model should perform perfectly if it truly looks at the image.
The Natural Color Dataset (NCD)~\cite{anwar2020image} is a dataset of various fruits colored either in the natural color or in gray.
Given a gray image of a fruit, we asked the model if the color is gray.
Surprisingly, the token likelihood score said no for many fruits and vegetables.

% It is likely that VL models have strong prior in color, [is BERT blind?] because most of images they learned would depict natural images(e.g Red apple), that would exist in real word, not learning unnatural images(e.g purple apple). To check whether our method can really reduce bias in color, we adopt NCD(Natural Color Dataset), which provides images of 20 types of fruits and vegetables paired with colors(also provide gray-colored version). \\
% Using same VL model(OFA) and differing only scoring method(TLC and ours), We feed 2 types of image-caption pair as input for model : ("The color of this {fruit} is gray", original image) and ("The color of this {fruit} is gray", gray-colored image),  then subtract the score of the former input from the score of the latter : score(caption, gray image) - score(caption, original image). Higher the difference, Less bias the model has, because high difference means that model can discriminate the color of fruit, more focusing on given visual information(fruit's gray color), not on language prior (fruit's original color). Figure 1 shows that using our method, comparing with TLC, VL model can concentrate more on visual information in all fruit type from NCD. \\[20pt]

\bfsection{Approach and metrics}
We use only the grayscale images from the NCD dataset, which leaves 723 images consisting of 20 different fruits.
Given an image, we build a true caption "The \textit{fruit} is gray." and an adversarial caption 
"The \textit{fruit} is \textit{color}.", in which \textit{fruit} and \textit{color} are variables replaced with the type of the fruit and its natural color.
%Given the grayscale image of a fruit or vegetable, we build the true caption "The \textit{fruit} is gray." and an adversarial caption "The \textit{fruit} is \textit{color}.".
Then, we compute the difference between the score of the true caption and that of the adversarial caption given the image.
Intuitively, the score should always be greater than zero for an image-text matching algorithm without severe language bias.

\bfsection{Results}
Fig~\ref{fig:color_bias} compares \modelname and token likelihood of the captioning model (TL) on the color debiasing experiment.
We report the ratio of the biased samples and the biased types (fruit or vegetable types of mean value lower than zero) here.
TL prefers its language bias over the image context in a third of the categories.
% On the other hand, TPMI, which alleviates the effect of language bias by suppressing the marginal likelihood, shows mean differences greater than zero for all categories.
In contrast, \modelname effectively reduces language bias, resulting in mean differences that are consistently above zero for all but one category.
% Note that while \modelname clearly reduces language bias, it cannot perfectly distinguish all biased samples.

\subsection{Debiasing Numbers}
\label{subsec:exp_number}

\begin{table*}[ht]
    \centering
    \begin{tabular}{l|ll|ccc|ccc}
        & & & \multicolumn{3}{c|}{Gender Bias $\downarrow$} & \multicolumn{3}{c}{Recall $\uparrow$} \\
        & Model & Method & Bias@1 & Bias@5 & Bias@10 & Recall@1 & Recall@5 & Recall@10 \\
        \hline
        \multirowcell{9}{T2I}
        &SCAN & ITC & 13.8 & 21.3 & 24.8 & 25.4 & 54.1 & 67.8 \\
        &FairSample & ITC & 11.3 & 19.2 & 22.9 & 26.8 & 55.3 & 68.5 \\
        &CLIP-clip & ITC & \textbf{6.7} & \textbf{14.7} & \textbf{16.1} & 27.3 & 50.8 & 62.0 \\
        \hhline{~--------}
        & CLIP
        & ITC & 8.8& 15.7& \underline{18.9} & 36.5& 61.0& 71.6 \\
        \hhline{~--------}
        &\multirowcell{3}{OFA\textsubscript{large}}
        & ITM & 13.2& 19.5& 23.3& 35.3& 67.1& 76.9 \\
        && TL & \underline{8.2}& \underline{15.4}& 19.2& 38.0& 65.1& 74.4 \\
        && \modelname & 9.2& 15.9& \underline{18.3}& \textbf{55.6}& \textbf{75.2}& \textbf{79.5} \\
        %\hhline{~--------}
        %& \multirowcell{2}{LLaVA} & TL & - & - & - & - & - & - \\ 
        %&  & MASS & - & - & - & - & - & - \\ 
        \hline
        \multirowcell{6}{I2T}
        & CLIP
        & ITC & 4.0& 11.4& 15.4& 56.9& 80.0& 87.0 \\
        % \cmidrule(lr){2-9}
        \hhline{~--------}
        &\multirowcell{3}{OFA\textsubscript{large}}
        & ITM & 6.3& 12.7& 15.7& 42.1& 79.7& 88.9 \\
        && TL & 6.6& 12.4& 15.3& 33.4& 70.7& 84.8 \\
        && \modelname & \textbf{3.5}& \textbf{9.9}& \textbf{13.4}& \textbf{57.3}& \textbf{84.7}& \textbf{89.8} \\
        %\hhline{~--------}
        %& \multirowcell{2}{LLaVA} & TL & - & - & - & - & - & - \\ 
        %&  & MASS & - & - & - & - & - & - \\ 
        \hline
        
    \end{tabular}
    \caption{Gender bias reduction experiment on MS-COCO test split. \textit{T2I} denotes text-to-image retrieval and \textit{I2T} denotes image-to-text. We compare our image-text matching score (\modelname) with the baselines on both the gender bias metric Bias@K and the retrieval metric Recall@K. The best numbers are \textbf{bolded} and the second-best numbers are \underline{underlined}.}
    \label{tab:gender}
\end{table*}

\iffalse
{'base': [0.3338888888888889, 0.7066666666666667, 0.8475],
 'tpmi': [0.5730555555555555, 0.8472222222222222, 0.8983333333333333],
 'tpmi_0.5': [0.22944444444444445, 0.6080555555555556, 0.8158333333333333],
 'itm': [0.42055555555555557, 0.7972222222222223, 0.8894444444444445],
 'clip': [0.5694444444444444, 0.7997222222222222, 0.8697222222222222]}
\fi

VL models are known to prefer specific numbers such as 0, 1, and 2~\cite{parcalabescu2021seeing},
regardless of the actual number of objects in an image.
Here, we test \modelname on the task of counting the number of visual entities.

%Thus, We additionally conduct an counting experiment to verify VL model can counts entities in image exactly beyond the training distribution, not distracted by prior incurred by imbalanced distribution in number. 

\bfsection{Approach}
We adopt the counting benchmark in VALSE dataset~\cite{parcalabescu2022valse}.
VALSE formulates counting tasks as foils. Given an image, a model is asked to differentiate the true caption with the correct number from the false caption with the wrong number (foil).
The benchmark consists of three sets: a set with balanced number distribution in the foils (\textit{balanced}), a set that only uses small numbers (\textit{small}), and a set in which true captions have only large numbers and foils have only small numbers (\textit{adversarial}). Note that models typically understand small numbers better since most images in training data do not have many objects. Hence, the \textit{adversarial} set is especially hard for models with language bias.
%in ~\cite{Parcalabescu2021VALSEAT}, which consists of various tasks that cover basic linguistic phenomena, we select tasks involving counting. They used Visual7W VQA dataset, and source its ‘how many’ examples whose answers are numerals. We followed experiment settings of ~\cite{Parcalabescu2021VALSEAT}.

\bfsection{Metric and baselines}
We report pairwise ranking accuracy metric, which formulates the counting task as an image-text alignment scoring problem.
Other baselines here include the multitask models such as LXMERT~\cite{tan2019lxmert} and 12-in-1~\cite{Lu_2020_CVPR}.

\bfsection{Results}
Table~\ref{tab:count} shows that our \modelname outperforms all baselines except one in all subsets.
Importantly, the score excels at the \textit{adversarial} set and performs on par with the best model 12-in-1.
As the \textit{adversarial} set is designed to be hard for models with language bias,
this result further verifies that the improvement of \modelname comes from language debiasing.

\subsection{Debiasing Genders}
\label{subsec:exp_gender}

\begin{figure}[t]
    \centering
    \includegraphics[width=0.48\textwidth]{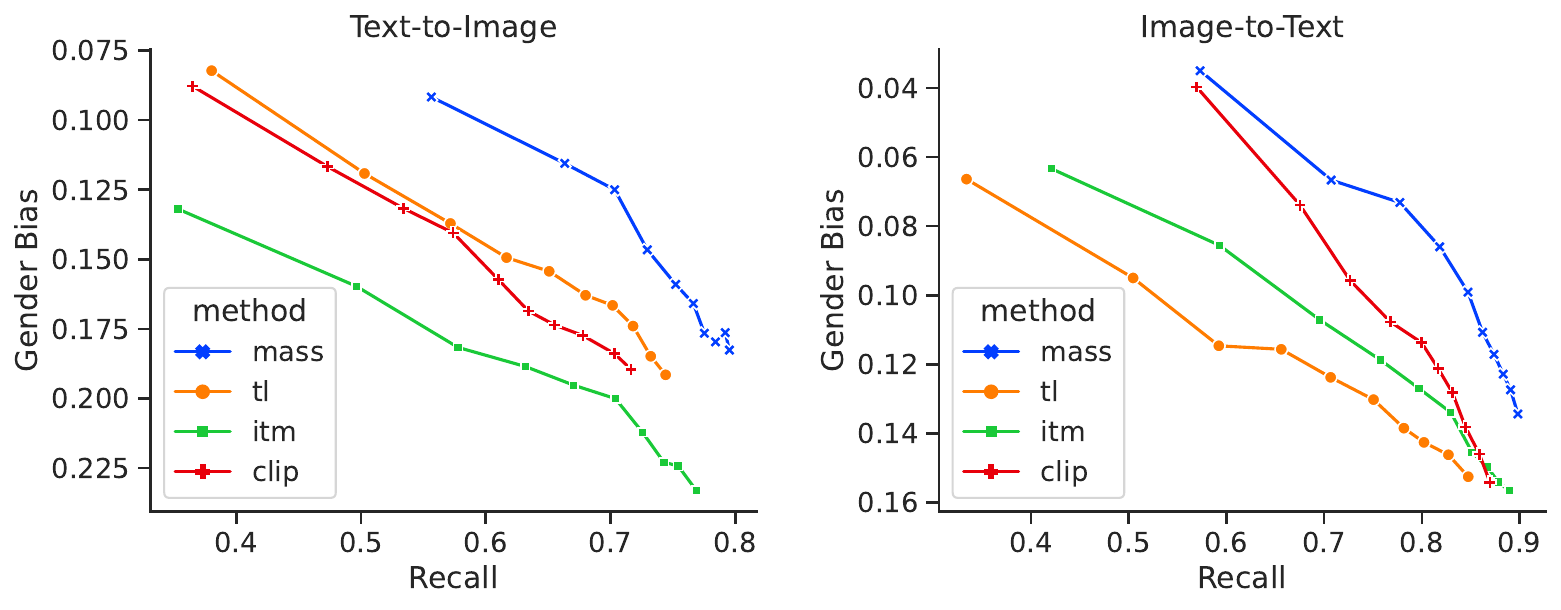}
    \caption{The Pareto frontier of recall-bias trade-off in COCO-captions. Y-axis (gender bias) is inverted for better visualization.}
    \label{fig:exp_gender_paretto}
\end{figure}

We now turn to a more practical problem of balancing gender bias in image-text matching.
Here, we adopt the conventional experiment setup~\cite{wang2021gender} to evaluate the gender bias of a text-based image search algorithm.
The hypothesis is simple:
given a gender-neutral text query, the retrieved images should contain a similar number of men and women in them. % for the search algorithm to be unbiased.

\bfsection{Approach}
We evaluate our method in both text-to-image and image-to-text retrieval tasks using the test split of MS COCO dataset~\cite{chen2015microsoft}.\footnote{We use the Karpathy split~\cite{karpathy2015deep} with 5000 test images.}
Our text-to-image experiment follows Wang~\etal's setup. First, we \textit{neutralize} each caption in the dataset by replacing any gender-specific word with the corresponding gender-neutral alternatives. Then, we perform text-to-image retrieval on the COCO dataset using neutralized captions.

For the image-to-text retrieval task, we follow a similar approach; however, neither captions nor images are neutralized to align with the definition of the bias score described below. Instead, retrieval is performed using original captions. Neutralization of images (e.g., masking gender-related features) was avoided as it led to a loss of critical contextual information necessary for valid retrieval comparisons.
We utilize CLIP~\cite{Radford2021LearningTV} to retrieve the top 20 candidates, followed by re-ranking the list using image-text matching algorithms. This two-stage retrieval strategy is commonly adopted to improve evaluation efficiency~\cite{wang2022git,li2023blip}.

For the gender debiasing experiments, we include additional baselines from prior debiasing methods: SCAN~\cite{lee2018stacked}, FairSample~\cite{lee2018stacked}, and CLIP-clip~\cite{wang2021gender}. It should be noted that our results for the CLIP model differ from those reported by Wang~\etal due to the use of a larger model scale (ViT-Large in our study versus ViT-Base in theirs).

\bfsection{Metrics}
The \textit{bias score} suggested in Wang~\etal~compares the proportions of masculine and feminine images or captions in the search results.
When $N_{m}$ and $N_{f}$ are respectively the number of retrieved images or captions depicting male and female in dataset $D$, the bias score for the top $K$ retrieved results ($Bias@K$) is defined as follows:

\begin{align}
    f(x) &:= \begin{cases}
        0 & \text{if~} N_{m} + N_{f} = 0 \\
      \frac{N_{m} - N_{f}}{N_{m} + N_{f}} & \text{otherwise}
    \end{cases} \\
    &Bias@K := \frac{1}{|D|} \sum_{x \in D} f(x)
\end{align}

Note that the query $x$ is an image to retrieve a list of captions in image-to-text retrieval, and vice versa for text-to-image.
We use the words in captions to decide whether an image or caption contains gender-specific information: refer to Wang~\etal~\cite{wang2021gender} for more details.

We also evaluate the standard \textit{recall} metric as an indicator of correctness: for each query, recall is one when the retrieved samples contain any of the ground-truth image-text pairs and zero in the other case.
A good algorithm should be able to minimize bias and keep the recall high.

\bfsection{Results}
Table~\ref{tab:gender} summarizes the results of our gender debiasing experiments.
In the text-to-image setting, \modelname outperforms all other baselines by a wide margin in recall, while maintaining the bias score close to the best non-debiasing method (TL) of the baselines.
Also, while a previous debiasing-specific method (CLIP-clip) effectively ameliorates gender biases, it comes at an expanse of high degradation of recall accuracy.

On image-to-text retrieval, the captions retrieved by \modelname contained much more balanced gender keywords than the baselines.
Further, \modelname still shows the best recall among others, proving that its social bias reduction does not come at an expanse of correctness.
In conclusion, \modelname is the best option here for gender-debiased retrieval,
as depicted in the Pareto curve of Figure~\ref{fig:exp_gender_paretto}.
\section{Experiments: Linguistic Complexity}
\label{sec:exp_comp}

\begin{table*}
    \centering
    \begin{tabular}{l|ccc|ccc|ccc}
    \small
         & \multicolumn{3}{c}{Text} & \multicolumn{3}{c}{Image} & \multicolumn{3}{c}{Group} \\
         Models
        & ITM/C & TL & \textbf{MASS}
        & ITM/C & TL & \textbf{MASS}
        & ITM/C & TL & \textbf{MASS}
        \\
        \hline
        Human & \multicolumn{3}{c}{89.5}  & \multicolumn{3}{c}{88.5}  & \multicolumn{3}{c}{85.5} \\
        Chance & \multicolumn{3}{c}{25.0} & \multicolumn{3}{c}{25.0} & \multicolumn{3}{c}{16.7} \\
        \hline
        VinVL & 37.8 & - & - & 17.8 & - & - & 14.5 & - & - \\
        LXMERT & 19.3 & - & - & 7.0 & - & - & 4.0 & - & - \\
        CLIP & 30.8 & - & - & 10.5 & - & - & 8.0 & - & - \\
        \hline
        OFA\textsubscript{large}
        & 26.4 & 26.8 & \textbf{32.0} ($\uparrow$5.2)
        & 16.5 & 29.8 & \textbf{31.5} ($\uparrow$2.7)
        & 9.5 & 15.8 & \textbf{20.3} ($\uparrow$4.5)\\
        \xspace\xspace
        \xspace\xspace\xspace\xspace\xspace\xspace\textsubscript{base}
        & 26.8 & 23.8 & \textbf{30.3} ($\uparrow$3.5)
        & 10.8 & \textbf{25.3} & 21.3 ($\downarrow$4.0)
        & 6.5 & 12.5 & \textbf{15.3} ($\uparrow$2.8) \\
        \xspace\xspace
        \xspace\xspace\xspace\xspace\xspace\xspace\textsubscript{tiny}
        & \textbf{22.8} & 15.5 & 18.0 ($\downarrow$4.8)
        & 7.8 & 14.8 & \textbf{17.5} ($\uparrow$2.7)
        & 4.5 & 5.0 & \textbf{8.0} ($\uparrow$3.0)\\
        \hline
        BLIP-2\textsubscript{T5}
        & \textbf{42.5} & 22.8 & 26.8 ($\downarrow$15.7)
        & 19.5 & 21.5 & \textbf{32.0} ($\uparrow$10.5)
        & 15.5 & 11.0 & \textbf{18.0} ($\uparrow$2.5) \\
        \xspace
        \xspace\xspace\xspace\xspace
        \xspace\xspace\xspace\xspace\xspace\textsubscript{OPT}
        & \textbf{42.5} & 23.3 & 22.8  ($\downarrow$19.7)
        & 19.5 & 24.8 & \textbf{29.8} ($\uparrow$5.1)
        & \textbf{15.5} & 12.5 & 15.0 ($\downarrow$0.5) \\
        \hline
        LLaVA & - & 27.0 & \textbf{32.3} ($\uparrow 5.3$) & - & 24.7 & \textbf{31.8} ($\uparrow 7.1$) &  - & 14.0 & \textbf{19.0} ($\uparrow 5.0$) \\
        \hline
    \end{tabular}
    \caption{Results on Winoground~\cite{thrush2022winoground}.
    The best matching algorithm for each model is marked \textbf{bold}. } % $\Delta$ denotes the improvement of \modelname over ITM.}
    \label{tab:comp}
\end{table*}
\begin{table*}
    \centering
    \begin{tabular}{l|ccc|ccc|ccc}
         & \multicolumn{3}{c}{Text} & \multicolumn{3}{c}{Image} & \multicolumn{3}{c}{Group} \\
         Models
        & ITM/C & TL & \textbf{MASS}
        & ITM/C & TL & \textbf{MASS}
        & ITM/C & TL & \textbf{MASS}
        \\
        \hline
        Chance & \multicolumn{3}{c}{25.0} & \multicolumn{3}{c}{25.0} & \multicolumn{3}{c}{16.7} \\
        \hline
        CLIP & 61.1 & - & - & 38.3 & - & - & 31.8 & - & - \\
        OFA\textsubscript{large}
        & 59.0 & 31.8 & \textbf{66.0} ($\uparrow$7.0)
        & 47.5 & 21.5 & \textbf{50.1} ($\uparrow$2.6)
        & 39.7 & 12.5 & \textbf{43.0} ($\uparrow$4.7) \\
        %LLaVA & - & - & - & - & - & - & - & - & - \\
        \hline
        %BLIP-2\textsubscript{T5}
        %& 77.0 & 25.5 & 53.5 ($\downarrow$23.5)
        %& 52.1 & 32.0 & 71.2 ($\uparrow$19.1)
        %& 48.8 & 14.3 & 48.0 ($\downarrow$0.8) \\
        %\xspace\xspace\xspace\xspace
        %\xspace\xspace\xspace\xspace\xspace\xspace\xspace\xspace\textsubscript{OPT}
        %& 77.0 & 23.3 & 54.4 ($\downarrow$18.0)
        %& 52.1 & 35.0 & 71.5 ($\uparrow$19.4)
        %& 48.8 & 14.3 & 48.6 ($\downarrow$0.2)\\
        %\hline
       
    \end{tabular}

      \caption{Result on SVO-Probes~\cite{hendricks2021probing}.
    %We composed 31,357 samples from 12,936 images (SVO-probe originally has 14,102 images, but some images are no longer accessible (the corresponding urls are corrupted).
    The best numbers are \textbf{bolded}.}
    \label{tab:svo-probe}
\end{table*}

As \modelname is built on captioning models, it could suffer less from the well-known linguistic insensitivity defect~\cite{yuksekgonul2022and} of ITC-based models such as CLIP~\cite{Radford2021LearningTV}.
However, we still need to verify whether the debiasing effect of \modelname also comes at a loss of this linguistic capability.
Thus we investigate whether \modelname can outperform the baselines in a visual-language compositionality benchmark
(\cref{subsec:exp_wino}) and a benchmark to test the capability to distinguish subject, noun, and verb in a caption (\cref{subsec:exp_svo}).
Both tasks require understanding linguistic structures beyond basic bag-of-words representations.

\subsection{Winoground}
\label{subsec:exp_wino}

The Winoground benchmark~\cite{thrush2022winoground} evaluates
a VL model's capability to understand compositionality.
A sample in Winoground consists of two images and two captions with two correct and two wrong image-text pairs.
Importantly, the two captions contain the same set of words but in a different order. This way, a model that perceives captions as bag-of-words~\cite{yuksekgonul2022and} without understanding visual-linguistic compositionality cannot do well on the task.
% is a meticulously chosen benchmark designed to assess models' capacity for compositional reasoning. The benchmark entails two images and two captions, and the objective is to correctly pair them together, despite the captions containing the same words in different sequences. \\

\bfsection{Metrics and baselines}
We use the suggested set of metrics~\cite{thrush2022winoground}: \textit{Textscore} which compares captions given the images, \textit{Imagescore} which compares images given the captions, and \textit{Groupscore} that is true only if both are true.
Note that \textit{Groupscore} is the preferred metric since
the other metrics can overestimate model performance by separating score computations for captions in the same sample~\cite{elazar2021back}. Baselines include popular VL models (VinVL~\cite{zhang2021vinvl}, LXMERT~\cite{tan2019lxmert}, and CLIP~\cite{Radford2021LearningTV}).

\bfsection{Results}
Table~\ref{tab:comp} shows \modelname outperforms all baselines by a wide margin in the main metric (\textit{Groupscore}). 
The same OFA\textsubscript{large} model saw more than twofold improvement by using \modelname over the ITM score.
Also, note that \modelname with OFA\textsubscript{large} is the only option that surpasses random chance in \textit{Groupscore}.
Finally, \modelname's contribution is better represented in \textit{Imagescore} than in \textit{Textscore},
indicating that \modelname can balance the unusually low \textit{Imagescore} of the baselines with the \textit{Textscore}.
% compares three method for scoring caption confidence: ITM, Token-Level score, TPMI(Token-level Pointwise Mutual Information, ours) on winoground benchmark. Noteworthy point in here is that previous method shows a significantly lower score in Imagescore compared to Textscore, which is a bottleneck for getting high score on Groupscore. Our methods, however, increases textscore usually, except one model size setting(OFA\textsubscript{tiny}), but increases Imagescore notably in every settings reported. With the huge increase in Imagescore, Groupscore also increased in all cases, compared with OFA ITM and OFA with TL (+11.8 on groupscore using OFA\textsubscript{large}, +8.3 with OFA\textsubscript{base}, and +4.8 with OFA\textsubscript{tiny}).

We also observe a similar trend of results with an alternative backbone model (BLIP-2~\cite{li2023blip}) as \modelname improves the base token likelihood method (TL) across all scores.
However, there are considerable gaps in Textscore between ITM and TL.
We suspect that the advanced training techniques used to optimize the ITM head in BLIP-2, such as hard negative mining~\cite{li2021align}, contributed to its better performance over the OFA model. Still, \modelname helps close the performance gap in the Groupscore.

%Table 3 shows groupscore in each subset, Core and Rest, and total set. Baseline methods(OFA ITM, OFA with TL scoring, and any other VL model) shows very slight difference between scores on Core and Rest. Using our method, however, score on Core subset gets more than doubles over score on Rest subset(with OFA\textsubscript{base}) and almost double(with OFA\textsubscript{large}, which indicates that estimating score with Pointwise Mutual Information enhances VL model's compisitional understanding. 
 
% \input{table/table_winoground_visualize}

\subsection{SVO-Probes}
\label{subsec:exp_svo}

SVO-Probes~\cite{hendricks2021probing} is another benchmark
testing VL models' sensitivity to linguistic alterations.
Here, the models are given a true caption and a false caption with 
either the subject, verb, or object changed to break the image-caption alignment.
% SVO-Probes dataset aims to check which type of word is hard for VL models to distinguish among subject, verb, and object.
% Even though Winoground~\cite{thrush2022winoground} is widely used to measure model's compositionality ~\cite{bugliarello2023measuring, petryk2023simple, pandey2022cross}, it is short of number of samples(consisting of 800 image-sentence pairs). So we opt for another benchmark, \textbf{SVO-Probe}~\cite{hendricks2021probing} because it offers a well-designed structure and a large amount of data (consisting of 421 verbs and over 48,000 image-sentence pairs). 

% The purpose of this benchmark is to test the comprehension of subject, verb, object in vision-language models, which is similar purpose of winoground: measuring compositionality. Each data instance in the dataset includes a textual caption, a positive image that matches the caption, and a controlled (adversarial) negative image. The negative image shares two aspects (subject, verb, or object) with the sentence but does not match the remaining aspect. 

\bfsection{Metrics and baselines}
Unlike the original dataset paper~\cite{hendricks2021probing},
we here use Winoground-style metrics for evaluation
to prevent overvaluing a VL model's response as in the previous section (\cref{subsec:exp_wino}).
% We also report the original metrics in~\cref{sec:ax_svo}.
Note that the original accuracy metric does not accept an arbitrary similarity function
since it requires a binary decision on the image-text pair match.
% Hence, we build \jw{detail}.
We include the vanilla CLIP model as the baseline here.

% Since previous researches have used SVO-probe dataset only to measure image-text matching with classification head ("does this sentence describe this image?"), we measure compositionality of VL model via measuring textscore, imagescore and groupscore, same as Table~\ref{tab:winoground}. We construct each sample to have 2 images and 2 captions, each captions describing each images, only differ in just one word. \\
% We test on SVO-probe dataset with same setting on winoground experiment. We additionally added CLIP as baseline backbone model for comparison.

\bfsection{Results}
As in the Winoground experiment,
Table~\ref{tab:svo-probe} also shows that \modelname 
can improve image-text matching in data with linguistic complexities.
\modelname outperforms all baselines on each metric,
showing that it distinguishes the linguistic aspects better than other methods.
%Also, the fine-grained results in~\cref{sec:ax_svo} reveal that the majority of \modelname's improvement comes from verb understanding, in which previous research showed that visual-language models tend to perform the worst.
%The results of experiment on SVO-probe can be observed in Table~\ref{tab:svo-probe}, which shows similar tendency as result on winoground(Table~\ref{tab:winoground}). TPMI outperforms baseline method by 4.9, 2.6, and 3.3 on textscore, imagescore, and groupscore respectively. 
% ~\cite{hendricks2021probing} claims that verbs in particular are an interesting challenge in image–language representation learning. To verify that our method can give capability of "Verb understanding" to backbone model, We additionally report result on three subsets which are divided based on whether changed word is a subject, verb or object.(\cref{sec:ax_exp_model}). 

\section{Related Work}

\bfsection{Visual-Language Model}
Early visual-language models like ViLBERT~\cite{lu2019vilbert}, VisualBERT~\cite{li2019visualbert}, and LXMERT~\cite{tan2019lxmert} established visual-language alignment based on pretrained text encoder representations.
Later, the success of CLIP~\cite{Radford2021LearningTV} introduced contrastive learning as a core method for visual-language models, leading to further research like ALBEF~\cite{li2021align}, which incorporates contrastive loss as a part of its multi-task losses.
BLIP-2~\cite{li2023blip} demonstrated that frozen large language models can enhance image captioning performance. OFA~\cite{wang2022ofa} and UnifiedIO~\cite{lu2022unified} further extend visual-language models to a wider range of tasks including image generation. Recent visual instruction tuning models allow prompt-based control of visually conditioned knowledge.
MiniGPT-4~\cite{zhu2023minigpt} and LLaVA~\cite{liu2024visual} extend the pretrained language model LLAMA~\cite{touvron2023llama} (and its variant Vicuna~\cite{chiang2023vicuna}) to multimodal inputs.
A prior study~\cite{lin2024revisiting} proposes a framework for mitigating language priors in vision-language models by introducing 'null images' and leveraging Pointwise Mutual Information (PMI), techniques similar to those presented in this work. Although both works share methodological elements, they were developed independently and address different objectives:~\cite{lin2024revisiting} focuses on theoretical contributions to vision-language alignment, while this work targets societal bias reduction and introduces new evaluation metrics, such as bias-aware Text-to-Image retrieval.

\bfsection{Bias in Multimodal Models}
Visual-language models often suffer from language bias induced by their training dataset. This deficiency has been explored as visual priming bias~\cite{zhang2016yin, goyal2017making} and language prior~\cite{agrawal2018don, ramakrishnan2018overcoming}.
Language bias is also closely connected to socially offensive bias,
such as gender bias~\cite{de2021stereotype} and racial bias~\cite{davidson2019racial} in language models.
Social biases are also observed in the VLMs~\cite{hendricks2018women, wang2021gender, zhao2021understanding, birhane2021multimodal}.
Other works explore diverse categories~\cite{garcia2023uncurated} and metric extension~\cite{ross2021measuring} for social bias understanding.

\iffalse
\bfsection{Compositional Reasoning}
Various benchmarks have been suggested to test visual-language models' capability to handle visual-linguistic compositionality, including single noun difference understanding~\cite{shekhar2017foil}, comprehension of subject-verb-object relationships~\cite{hendricks2021probing}, sensitivity to the linguistic property perturbations~\cite{parcalabescu2022valse}. 
Aside from investigating linguistic properties, other works also explored
comparison of common attributes observed across multiple images~\cite{bogin2021covr} and identification of the difference between hard-negative images~\cite{thrush2022winoground}.
Recently, visual-language models such as CLIP have turned out to show weak language modeling capability~\cite{yuksekgonul2022and}.
In response, new benchmarks such as CREPE~\cite{ma2023crepe} and SugarCrepe~\cite{hsieh2024sugarcrepe} are proposed to measure the visual-linguistic compositionality of the models without incorporating linguistic bias.
\fi
\section{Conclusion}
\label{sec:conclusion}

We introduce \modelname (\modelnamelong) as an inference-time framework to address the issue of language bias in image-text matching.
% \modelname leverages pretrained visual-language models to extract image-conditional and text-only likelihoods per text token by computing pointwise mutual information between the image and each text token and aggregating them to generate a scalar similarity score.
\modelname builds on image captioning models to extract image-text matching capability devoid of their linguistic biases.
Specifically, it computes pointwise mutual information between the image and each text token and aggregates them to generate a scalar similarity score.
The proposed method significantly reduces language bias without additional training and demonstrates improved performance in debiasing tasks and visuo-linguistic compositionality tests, outperforming existing models such as CLIP and token-level likelihood.

We hope that our research sparks interest in
1. investigating linguistic bias inherent in the image-text alignment mechanism of the recent visual-language models
and 2. devising a method to reduce such language bias in off-the-shelf visual-language models without incorporating the computationally overwhelming training process.
\section*{Broader Impact}
\label{sec:limitations}

This paper introduces \modelname as a training-less algorithm
for reducing language bias in visual-language models.
As we point out in the experiments~(\cref{subsec:exp_gender})
our bias reduction framework also is closely related to social bias
in the VL models.
Thus, \modelname can be used to promote fairness,
for example, in the standard multimodal search setting
or in the evaluation of model generation results with regard to social or language bias.

However, the social bias reduction effect of \modelname has two important limitations.
First, it relies on the output of visual-language models, which may encode hidden biases.
Since \modelname is an inference-time method, it cannot audit the information stored in the models' parameters.
Hence, we acknowledge that while our framework can reduce social bias, the range and intensity of its effect
may vary with respect to the base VL model used.
Second, we only explore social bias in terms of language bias.
Social bias could be manifested as visual bias
or even in the form of bias that is identified only as combinations of image and text~\cite{yamada2022lemons}.
We hope this research serves as a simple but effective baseline 
to initiate a search for more robust approaches with regard to social bias reduction 
in image-to-text alignment problems.
\section*{Acknowledgements}
\label{sec:ack}

This work was partly supported by an IITP grant funded by the Korean Government (MSIT) (No.RS-2020-II201361, Artificial Intelligence Graduate
School Program (Yonsei University) and RS-2024-00353131) and the National Research Foundation of Korea (NRF) grant funded by the Korea government (MSIT) (No. RS-2024-00354218).

\bibliography{custom}

\clearpage
\appendix

% \section*{Appendix}\label{sec:appendix}
% \label{sec:ax_exp_model}

\section{Overview}
\label{sec:ax_overview}

We provide the following details in this appendix:
\begin{itemize}
    % \item In \cref{sec:ax_wino_backbone} we evaluate \modelname with an alternative base VL model.
    \item In \cref{sec:ax_wino_tag} we divide our Winoground experiment into fine-grained categories for the analysis of results.
    \item In \cref{sec:ax_svo} we revisit our SVO-Probes experiment with a fine-grained categorization to which of subject, verb, and object benefits from \modelname the most.
    \item In  \cref{sec:ax_implementation} we provide implementation details, including model weights and computational requirements.
    \item In \cref{sec:ax_samples} we provide samples from our gender bias and Winoground experiments.

\end{itemize}

\section{A Closer Look at the Winoground Results}
\label{sec:ax_wino_tag}

A recent work~\cite{diwan2022winoground} argues that not all samples in Winoground are appropriate for evaluating compositionality.
Some examples require detecting microscopic objects, which is more about detection capability than compositional understanding. Not only that, some samples require other capabilities,
which hinder an accurate evaluation of visual-linguistic compositionality. ~\cite{diwan2022winoground} annotated them as Non-compositional, Ambiguously Correct, Visually Difficult, Unusual Text, Complex Reasoning, and Unusual Image.
Samples with no external complexities are marked No-Tag.

% denotes any hindrance not existing in the sample, which is suitable for pure compositionality measurement.

We show the full results of each category in Table~\ref{tab:ax_winoground_tag}.
For a better interpretation of these results,
we summarize our findings in Table~\ref{tab:ax_winoground_core_rest}.
According to the table, most of the \modelname's improvement comes from the No-Tag subset. 
Since No-Tag is the subset without any complexities unrelated to compositionality,
we conclude that \modelname assists the compositional understanding of VL models.

% \input{table/ax_table_winoground_blip}
% To see whether our method truly enhances a compositional understanding of the backbone model, we divide Winoground dataset into two subsets and evaluate scores on each: 'Core' subset consists of samples that ~\cite{diwan2022winoground} tagged as No-tag (referred to as vanilla Winoground examples), and 'Rest' which consists of rest of examples in 'Core' subset. Among 400 samples in Winoground, 172 are included in Core subset, and the other 228 are in Rest subset. Table~\ref{tab:winoground_core_rest} shows the result on each subset. PMILD scoring enhanced \textit{Groupscore} on Core subset, a better environment for measuring compositionality solely, while no enhancement on Rest subset. This result indicates that applying pointwise mutual information score for language debiasing leads to improvement in compositional understanding of the VL model, not on other capabilities such as reasoning or object detection.

%\input{table/table_svo-probe_subset}

%\input{table/table_multimodal-predicate-noun-dependencies}
\section{A Closer Look at the SVO-Probes Results}
\label{sec:ax_svo}

The SVO-Probes benchmark~\cite{hendricks2021probing}
is originally proposed to check which among subject, verb, and object VL models find hard to understand.
Following this setting, we divide the data into Subject, Verb, and Object subsets 
to test how \modelname fares against each linguistic challenge.
Specifically, the Subject subset only consists of true captions and false captions with subject modification, and the Verb and Object subsets are defined likewise.
% suggests that VL models tend to show undesirable results in verb understanding compared with subject or object. 
% Since every caption pairs of SVO-Probes samples are one of three cases: to make an adversarial caption, which word in a sentence is modified among subject, verb, and object? Based on it, we make a subset by subject, verb, and object separately and show each subset's results reported in Table ~\ref{tab:svo-probe}.

%Table~\ref{tab:svo-probe_svoanalyze} tells that \modelname improves TL greatly, but its score difference with ITM varies depending on the subset.
Table \ref{tab:svo-probe_svoanalyze} indicates that \modelname significantly enhances token likelihood (TL), but the difference in scores compared to ITM varies across different subsets.
Importantly, \modelname shows clear enhancement in verb understanding, 
which is the task VL model malfunctions the most according to previous research~\cite{hendricks2021probing}.
However, \modelname also falls behind ITM by a considerable margin in the Object subset, leaving room for further improvement.
% Although we have not completely solved degeneration on verb understanding since \textit{groupscore} on verb subset(37.2) is lower than on subject or object subset(49.9 and 55.3), our method chiefly increases the performance on verb understanding, making the gap close between results on subject or object. Using PMILD, the VL model focuses more on the image by reducing linguistic bias, enhancing the backbone model's verb understanding, which has been a bottleneck for overall performance for multimodal compositional understanding.

\begin{table}[t]
    \centering
    \begin{tabular}{ll|ccc}
        \multicolumn{2}{c}{Models} & No-Tag & Rest & Full \\
        \hline
        \multicolumn{2}{l|}{Random Chance} & 16.7 & 16.7 & 16.7 \\
        \hline
        LXMERT & ITM & 4.1 & 3.9 & 4.0 \\
        UNITER & ITM & 7.0 & 13.1 & 10.5 \\
        CLIP & ITC & 8.2 & 7.9 & 8.0 \\
        \hline

        \multirowcell{4}{OFA\textsubscript{large}}
         & ITM & 10.5 & 8.8 & 9.5 \\
         & TL & 15.7 & \textbf{15.8} & 15.8 \\
        % & WTPMI & 20.3 & 16.2 & 18.0 \\
         & \modelname & \textbf{26.7} & 15.4 & \textbf{20.3} \\
         & $\Delta$  & +11.0 & -0.4 & +4.5 \\
         \hline
                 
        \multirowcell{4}{OFA\textsubscript{base}}
         & ITM & 8.7 & 6.6 & 7.5 \\
         & TL & 15.1 & 10.6 & 12.5 \\
        % OFA\textsubscript{base} & WTPMI & 18.6 & 12.3 & \textbf{15.0} \\
         & \modelname & \textbf{18.6} & \textbf{12.8} & \textbf{15.3} \\
         & $\Delta$  & +3.5 & +2.2 & +2.8 \\
         \hline
         
         \hline

    \end{tabular}
    \caption{
    \textit{No-Tag} denotes a subset that focuses on compositionality
    and \textit{Rest} is a subset with other unrelated complexities.
    \textit{Full} results are the same as the main Winoground experiment.
    $\Delta$ is the difference between \modelname and TL.
    The best numbers in each subset are \textbf{bolded}.
    We only report \textit{Groupscore} here.
    }
    \label{tab:ax_winoground_core_rest}
\end{table}

\begin{table*}[t]
\setlength{\tabcolsep}{2pt}
    \centering
    \tiny
    % \multicolumn{3}{c|}{\parbox{1.8cm}{\centering Ambiguously Correct}}
    \begin{tabular}{ll|ccc|ccc|ccc|ccc|ccc|ccc|ccc|ccc}

& & \multicolumn{3}{c|}{\parbox{1.6cm}{\centering Non Compositional}} & \multicolumn{3}{c|}{\parbox{1.6cm}{\centering Ambiguously Correct}} & \multicolumn{3}{c|}{\parbox{1.6cm}{\centering Visually Difficult}} & \multicolumn{3}{c|}{\parbox{1.6cm}{\centering Unusual Text}} & \multicolumn{3}{c|}{\parbox{1.6cm}{\centering Complex Reasoning}} & \multicolumn{3}{c|}{\parbox{1.6cm}{\centering Unusual Image}} & \multicolumn{3}{c|}{\parbox{1.6cm}{\centering NoTag}} & \multicolumn{3}{c}{\parbox{1.6cm}{\centering Full}}\\
\multicolumn{2}{c|}{Models} & T & I & G & T & I & G & T & I & G & T & I & G & T & I & G & T & I & G & T & I & G & T & I & G\\
\hline
CLIP & ITM & 76.6 & 36.7 & 33.3 & 30.4 & 15.2 & 13.0 & 15.8 & 0.0 & 0.0 & 30.0 & 16.0 & 10.0 & 24.4 & 7.7 & 3.9 & 25.00 & 9.0 & 5.4 & 30.4 & 11.1 & 8.2 & 30.75 & 10.50 & 8.00\\

LXMERT & ITM & 10.0 &13.3& 3.3& 10.9 &2.2& 0.0& 21.1 & 7.9& 2.6& 10.0 & 6.0& 0.0& 16.7 & 3.9& 1.3& 12.5 & 7.1& 1.8& 19.9 & 4.7& 4.1 & 19.25 & 7.00 & 4.00\\ 

\hline

 & ITM & 46.7 & 36.7 & 26.7 & 28.3 & 17.4 & 13.0 & 23.7 & 13.2 & 5.3 & 22.0 & 16.0 & 10.0 & 16.7 & 9.0 & 1.3 & 16.1 & 7.1 & 7.1 & 30.8 & 19.2 & 10.5 & 26.3 & 16.5 & 9.5\\
 
OFA\textsubscript{large} & TL & 36.7 & 46.7 & 23.3 & 37.0 & 21.7 & \textbf{15.2} & 18.4 & 21.1 & 18.4 & 22.0 & 34.0 & 14.0 & 16.7 & 24.4 & \textbf{9.0} & 30.4 & 35.7 & \textbf{21.4} & 27.9 & 29.7 & 15.7 & 26.8 & 29.8 & 15.8\\
 
 & \modelname & 43.3 & 46.7 & \textbf{33.3} & 23.9 & 26.1 & 13.0 & 36.8 & 26.3 & \textbf{21.1} & 30.0 & 26.0 & \textbf{18.0} & 19.2 & 15.4 & 6.4 & 25.0 & 26.8 & 17.9 & 38.4 & 39.5 & \textbf{26.7} & 32.0 & 31.5 & \textbf{20.3}\\
 \hline

& ITM & 33.3& 16.7& 10.0&21.7& 8.7& 4.4&15.8& 15.8& 10.5&24.0& 22.0& \textbf{12.0}&19.2& 3.9& 3.9&17.9& 9.0& 3.6&27.3& 12.8& 8.7&24.8& 11.8& 7.5\\
OFA\textsubscript{base} & TL & 46.7& 43.3& \textbf{26.7}&15.2& 19.6& 6.5&26.3& 18.4& \textbf{15.8}&18.0& 22.0& \textbf{12.0}&14.1& 14.1& \textbf{5.1}&35.7& 25.0& \textbf{16.1}&27.3& 31.4& 15.1&23.8& 25.3& 12.5\\
 & \modelname & 50.0& 30.0& \textbf{26.7}&32.6& 23.9& \textbf{21.7}&26.3& 21.1& 13.2&18.0& 14.0& 4.0&16.7& 11.5& \textbf{5.1}&28.6& 19.6& 14.3&36.6& 25.0& \textbf{18.6}&30.3& 21.3& \textbf{15.3}\\

 \hline
    \end{tabular}
    \caption{Results on Winoground~\cite{thrush2022winoground} benchmark categorized by reasons of difficulty~\cite{diwan2022winoground}. T, I, G denote \textit{Text}, \textit{Image}, and \textit{Groupscore}, respectively. The best numbers are \textbf{bolded} in \textit{Groupscore}.}  % result of LXMERT follows ~\cite{diwan2022winoground}}
    \label{tab:ax_winoground_tag}

\setlength{\tabcolsep}{6pt}

\end{table*}

%%% DATA
% OFA-L TPMI
% only_pmi {'ok': ([0.3953488372093023, 0.37209302325581395, 0.26744186046511625], 172), 'all': ([0.315, 0.32, 0.205], 400), 'Ambiguously Correct': ([0.2608695652173913, 0.2826086956521739, 0.13043478260869565], 46), 'Visually Difficult': ([0.2631578947368421, 0.3684210526315789, 0.21052631578947367], 38), 'Unusual Text': ([0.26, 0.3, 0.18], 50), 'Complex Reasoning': ([0.15384615384615385, 0.19230769230769232, 0.0641025641025641], 78), 'Unusual Image': ([0.26785714285714285, 0.25, 0.19642857142857142], 56), 'Non Minimal': ([0.4666666666666667, 0.36666666666666664, 0.3333333333333333], 30)}

% OFA-B TPMI
% only_pmi {'ok': ([0.2616279069767442, 0.37790697674418605, 0.20348837209302326], 172), 'all': ([0.21, 0.3175, 0.15], 400), 'Ambiguously Correct': ([0.15217391304347827, 0.30434782608695654, 0.08695652173913043], 46), 'Visually Difficult': ([0.13157894736842105, 0.2631578947368421, 0.07894736842105263], 38), 'Unusual Text': ([0.14, 0.28, 0.08], 50), 'Complex Reasoning': ([0.0641025641025641, 0.20512820512820512, 0.05128205128205128], 78), 'Unusual Image': ([0.25, 0.30357142857142855, 0.14285714285714285], 56), 'Non Minimal': ([0.3333333333333333, 0.4666666666666667, 0.23333333333333334], 30)}
\begin{table*}[t]
    \centering
    \begin{tabular}{ll|ccc|ccc|ccc}
         & & \multicolumn{3}{c|}{Subject (4938)} & \multicolumn{3}{c}{Verb (19918)} & \multicolumn{3}{c}{Object (6501)}\\
         \hline
         Model & Method & text & image & group & text & image & group &text & image & group  \\
        \hline
        
         CLIP\textsubscript{base}
        & ITC & \underline{66.1} & 45.7 & 38.6 & \underline{54.0} & 35.3& 28.0 & 79.0 & 41.6 & 38.3\\
         %CLIP\textsubscript{large}
        %& ITC & \textbf{77.2}& 56.1& \textbf{51.8}& 53.2& 34.5& 27.0 & 76.4& 41.9& 37.8\\
 
        %\hhline{~--------}
        \hline
        \multirowcell{3}{OFA\textsubscript{large}}
        & ITM & 64.5& \underline{55.5}& \underline{47.7}& 50.7& \underline{39.0}& \underline{30.0} & \underline{80.0}& \textbf{67.4}& \textbf{63.2}\\
        & TL & 49.4& 35.4& 27.4& 18.1& 17.3& 6.8 & 60.5& 23.7& 18.7\\
        & \modelname & \textbf{68.1}& \textbf{57.3}& \textbf{49.9}& \textbf{60.6}& \textbf{45.6}& \textbf{37.2} & \textbf{81.1}& \underline{58.6}& \underline{55.3}\\
        %& $\Delta$ & +2.0 & +1.8 & +2.2& +6.6 & +6.6 & +7.2 & +1.1 & -8.8& -7.9\\
        \hline

    \end{tabular}
    \caption{SVO-Probes~\cite{hendricks2021probing} results categorized by Subject, Verb, and Object modification.
    Each subset has a different number of samples notated alongside the subset name.
    %The number denoted in the bracket is the number of samples in each subset.
    The best numbers are \textbf{bolded} and the second best ones are \underline{underlined}.}
    % $\Delta$ denotes the score difference between PMILD and the best among baselines.}
    \label{tab:svo-probe_svoanalyze}
\end{table*}

\section{Implementation Details}
\label{sec:ax_implementation}
\bfsection{Backbone VL Models}
% We use three different sizes of OFA~\cite{wang2022ofa}, and their parameter counts are 930M for OFA\textsubscript{large}, 180M for OFA\textsubscript{base}, and 33M for OFA\textsubscript{tiny} each.
We adopt three versions of the base VL model, OFA~\cite{wang2022ofa};
\textit{large} (930M parameters), \textit{base} (180M), and \textit{tiny} (33M).
Following previous research~\cite{petryk2024simple}, we use 
the COCO-caption finetuned checkpoint when available (large), and use the base pretrained weights otherwise (base and tiny)
\footnote{The weights are publicly available at \url{https://github.com/OFA-Sys/OFA/blob/main/checkpoints.md}.}.
For the CLIP~\cite{Radford2021LearningTV} baseline, we use the \textit{base} version (86M parameters)
\footnote{Available at \url{https://github.com/openai/CLIP}.}.

\bfsection{Computational Requirements}
We use a single NVIDIA 3090Ti GPU (24GB Memory) for all experiments.
% Since our method involves inference(language modeling) without model training, we only report running time for inference. 
A forward pass of OFA\textsubscript{large} takes 0.27 seconds in our environment.
\modelname requires two forward passes to calculate pointwise mutual information,
which yields an overall runtime of 0.55 seconds (1.8it/s) per sample.
%Using OFA, it takes 0.27 second for one forward pass in inference; thus 0.55 second(1.8it/s) is required per one sample since we need two forward passes to calculate pointwise mutual information.

\bfsection{Dataset Sources}
%we use 2 dataset in this paper: Winoground~\cite{thrush2022winoground} and SVO-probe~\cite{hendricks2021probing}.
%For Winoground, we used original data~\footnote{https://huggingface.co/datasets/facebook/winoground} without any modification. There are 800 distinct images along with corresponding captions. Thus we evaluate on 400 instances, each consisting of two pairs of images and captions.
Here, we describe the sources of each data we use in the experiments;
Natural Colors Dataset~\cite{anwar2020image}~\footnote{\url{https://github.com/saeed-anwar/ColorSurvey}},
VALSE~\cite{parcalabescu2022valse}~\footnote{\url{https://github.com/Heidelberg-NLP/VALSE}},
Winoground~\cite{thrush2022winoground}~\footnote{\url{https://huggingface.co/datasets/facebook/winoground}},
and SVO-Probes~\cite{hendricks2021probing}~\footnote{\url{https://github.com/deepmind/svo_probes}}.

There are three additional points to note.
First, our gender bias experiments use images from COCO Captions~\cite{chen2015microsoft} dataset.
Also, the required gender-neutral captions are resourced from previous research~\cite{wang2021gender}
~\footnote{\url{https://github.com/eric-ai-lab/Mitigate-Gender-Bias-in-Image-Search}}.
Second, we could only obtain a subset of images for the SVO-Probes dataset due to some links being no longer accessible.
We download 12,906 among the full 14,102 images and conduct experiments with them.
%For SVO-Probes, we downloaded images from the official repository. \footnote{https://github.com/deepmind/svo_probes} SVO-Probes consists of 14,102 images and 36,842 samples by changing the subject, verb, or object from the original caption. However, since some image URLs are not accessible now, We only use 12,906 images and 31,356 samples for evaluation.
Finally, when evaluating SVO-Probes using the Winoground-style metrics,
we build false captions using the ground-truth alternative term annotations and the Spacy parser~\cite{honnibal2017spacy}.

\bfsection{Preprocessing}
We follow the preprocessing pipelines of the base VL models.
% For every backbone model we use, we follow preprocessing for images and text suggested in their official implementations. In the experiment on SVO-Probes, since they do not provide negative sentence(differ in only one word from the positive sentence) itself, instead providing which word is changed. So we manually make negative sentences for each sample using Spacy.

\bfsection{Hyperparameter Search}
We did not perform any hyperparameter search.
% No hyperparameter search is applicable to our inference-time framework, which involves no additional training.
%As our method does not involve any training process, no hyperparameter search is not carried out. 

\bfsection{Randomness}
Our method and all the baselines are deterministic functions given the pretrained models.
Hence, no randomness analysis is applicable to our research.
% As our method is a deterministic confidence-scoring algorithm, no randomness is involved in any of the experiments. Because SVO-probe provides image URLs instead of image files, variance on experiment using SVO-probe might occur.

\section{Qualitative Samples}
\label{sec:ax_samples}

\begin{figure*}[t]
    \centering
    \includegraphics[width=0.98\textwidth]{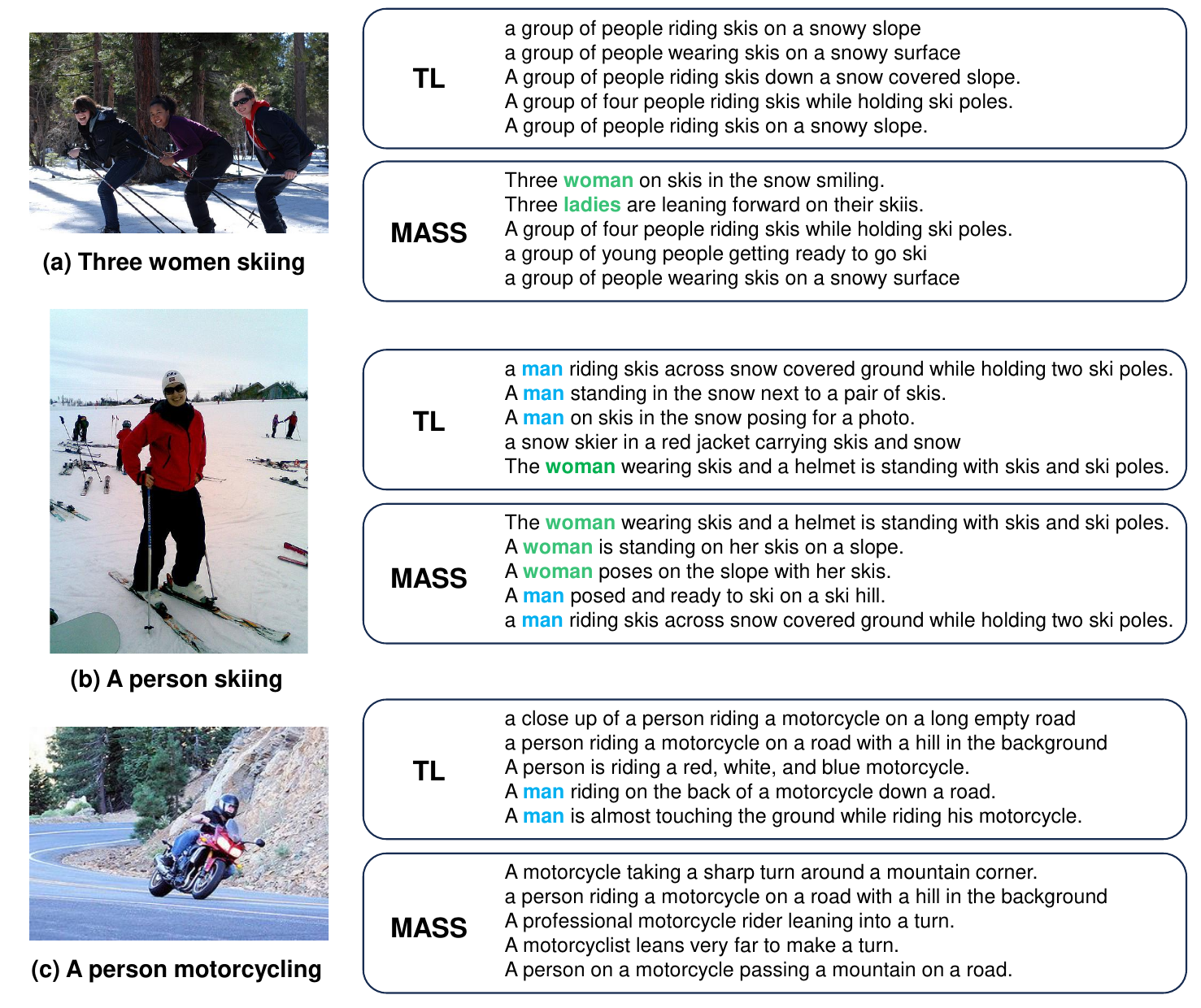}
    \caption{Top 5 COCO Captions image-to-text retrieval results, sorted by decreasing retrieval score. The token likelihood score (TL) avoids associating female keywords with sports activities, 
    by refusing to retrieve captions with clearly correct gender information in (a),
    or preferring to classify the skier as a man when the visual features possibly indicate otherwise (b),
    or hallucinate the gender information of the rider not visible in the image in (c).
    \modelname reduces such gender bias in its retrieved caption examples.}
    \label{fig:ax_samples_gender}
\end{figure*}

\begin{figure*}[t]
    \centering
    \includegraphics[width=0.98\textwidth]{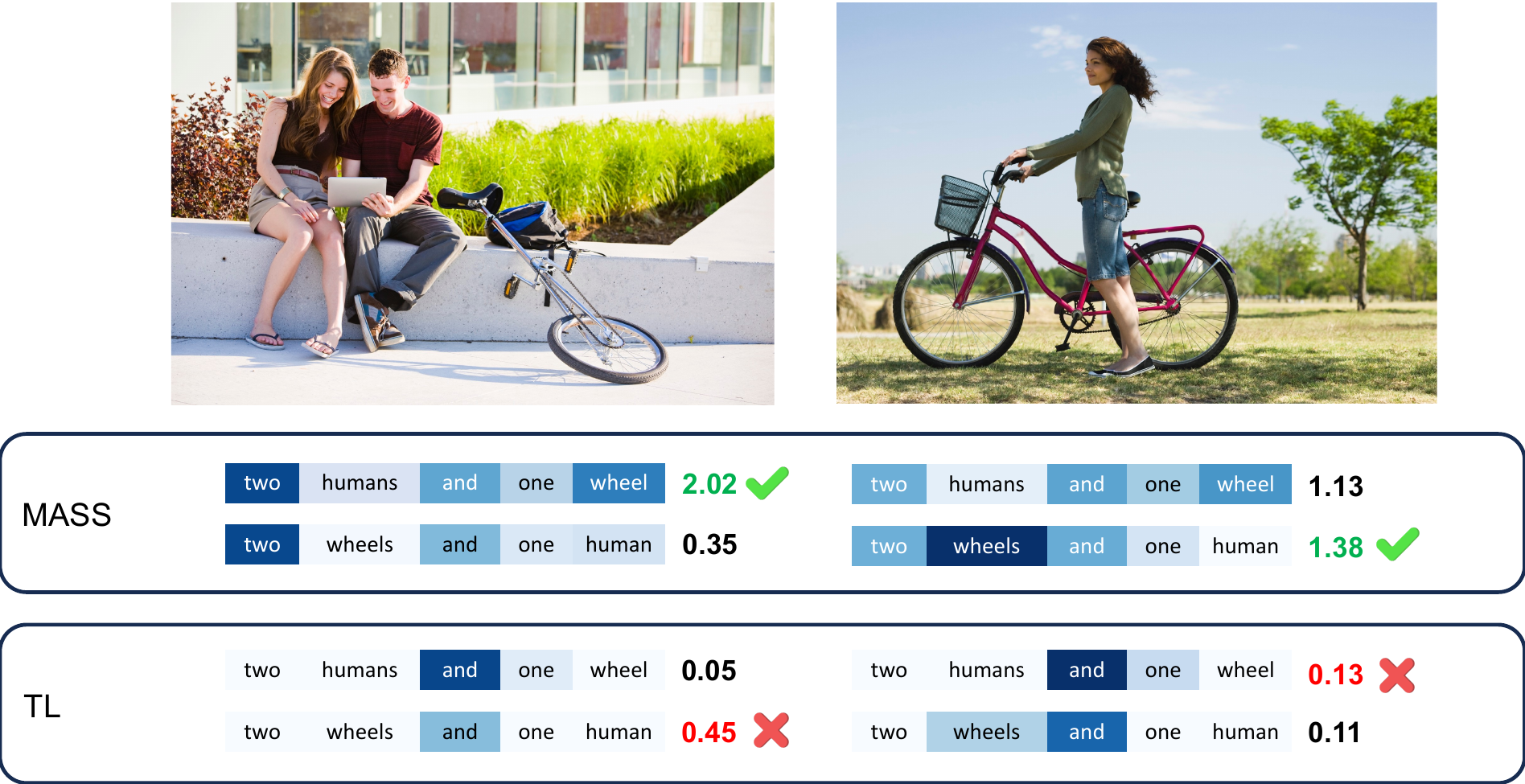}
    \caption{
    A qualitative sample from our Winoground~\cite{thrush2022winoground} experiment.
    The ground-truth image-text pairs are marked with $\checkmark$.
    \modelname assigns different scores per token depending on the image,
    leading to the correct judgment of image-text similarity.
    \modelname focuses more on quantifiers and objects,
    which are visually grounded words required for accurate image-text matching.
    On the other hand, TL focuses on the non-visual word \textit{and} regardless of the image and caption
    to yield wrong answers.
    }
    %An example of Winoground, comparing the result of each scoring method(TL and PMILD). The intensity of color in each caption means how much VL model scores to each word(token), and caption annotated with a checkmark means matched sentence with the above image. token likelihood score (TL) scores mostly on the word("and"), not focusing on the objects("human", "wheel") or quantifiers("one", "two"). In contrast, PMILD mainly scores more on quantifiers and objects, which leads to success in matching each caption to corresponding images, whereas TL fails both.}
    \label{fig:ax_sample_winoground_success}
\end{figure*}

\begin{figure*}[t]
    \centering
    \includegraphics[width=0.98\textwidth]{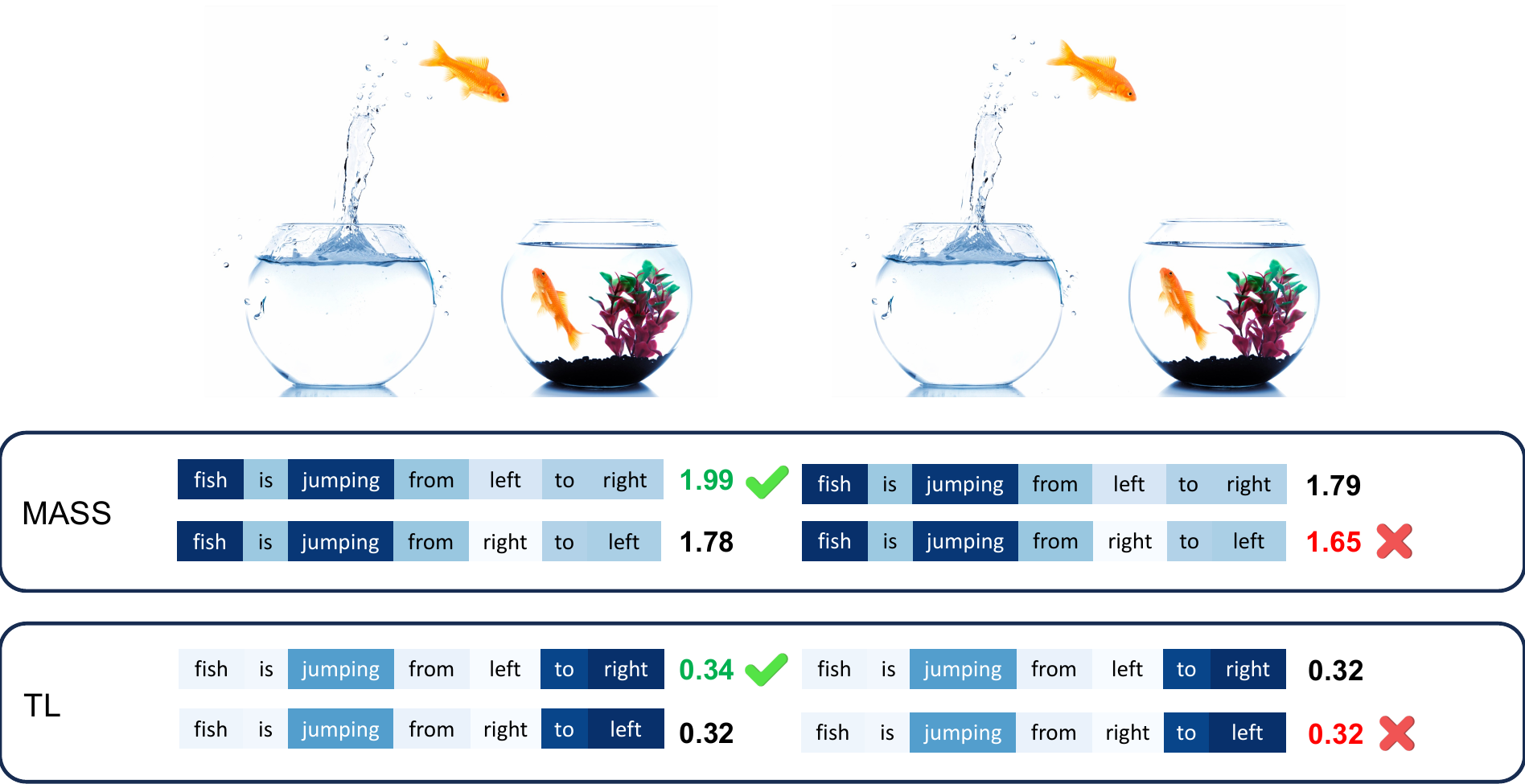}
    \caption{failure case analysis}
    \label{fig:ax_sample_winoground_failure}
\end{figure*}

% Here, we provide qualitative samples from our gender bias reduction (Figure~\ref{fig:ax_samples_gender}) and Winoground (Figure~\ref{fig:ax_sample_winoground}) experiments. Refer to the figure captions for the interpretation of each result.
Here, we provide two qualitative samples in our gender bias reduction (Figure~\ref{fig:ax_samples_gender}) and Winoground (Figure~\ref{fig:ax_sample_winoground_success}) experiments, and a failure case in Winoground (Figure~\ref{fig:ax_sample_winoground_failure}) experiment. Refer to the corresponding figure captions for the interpretation of each result.

\end{document}